\pdfoutput=1

\documentclass{article}

\usepackage{microtype}
\usepackage{graphicx}
\usepackage{subfigure}
\usepackage{booktabs}
\usepackage{xcolor}
\usepackage{hyperref}

\usepackage[accepted]{mlsys2027}
\makeatletter\renewcommand{\Notice@String}{Preprint.}\makeatother
\AtBeginDocument{\setlength{\footskip}{24pt}}
\fancypagestyle{plain}{\fancyhf{}\fancyfoot[C]{\thepage}\renewcommand{\headrulewidth}{0pt}}
\usepackage{placeins}

\newcommand{\methodnameshort}{\ensuremath{\mathrm{SH}_q}}

\mlsystitlerunning{\texorpdfstring{\methodnameshort}{SHq}: Learnable Grids for LLM Weight Quantization}

\begin{document}

\twocolumn[
\mlsystitle{\texorpdfstring{{\boldmath\methodnameshort}}{SHq}: Learnable Grids for LLM Weight Quantization}

\begin{mlsysauthorlist}
\mlsysauthor{Michael Helcig}{ethz}
\mlsysauthor{Earl Killian}{rhoa}
\mlsysauthor{Dan Alistarh}{ista}
\end{mlsysauthorlist}

\mlsysaffiliation{ethz}{Department of Computer Science, ETH Zürich, Zurich, Switzerland}
\mlsysaffiliation{ista}{Institute of Science and Technology Austria, Klosterneuburg, Austria}
\mlsysaffiliation{rhoa}{Rhoa Research, CA, USA}

\mlsyscorrespondingauthor{Michael Helcig}{mhelcig@ethz.ch}

\mlsyskeywords{post-training quantization, codebook, lookup table, LLM inference}

\vskip 0.3in

\begin{abstract}
Block-scaled lookup-table (LUT) formats, which represent weights as indices into a small value set (the grid), have become standard for low-bit LLM quantization, ranging from NF4 and BOF4 to the 3-bit LUT instruction in NVIDIA Rubin. While these formats leave the grid definition flexible, existing implementations rely on fixed grids derived either from numerical data types (INT, FP) or assumed weight distributions (NF, BOF). We demonstrate that no single fixed grid is universally optimal; the best fixed grid varies with bit-width: BOFS at 2 bits, NF at 4 and 5 bits, and INT at 6 bits. We introduce \methodnameshort, a closed-form family of grids parameterized by two critical degrees of freedom: a companding strength $\alpha$ that interpolates level spacing from uniform to floating-point-like, and an offset $c$ that introduces asymmetry. This offset allows the sign bit of the FP8 block scale, otherwise unused, to select between the grid and its mirror image, up to doubling the number of distinct values available per block. \methodnameshort{} modifies only the lookup table contents and serves as a drop-in replacement for the grid in any such format. Across five models ranging from 3B to 35B parameters, including a mixture-of-experts architecture, evaluated at 2--6 bits using RTN and GPTQ, \methodnameshort{} achieves the lowest KL divergence in every setting at 3 and 4 bits and in most settings at 2, 5 and 6 bits. It outperforms the best fixed grid by up to 34\% and remains superior under mixed-precision configurations. At 4.5 bits per weight, the \methodnameshort{} format reduces the KL divergence of NVFP4 by 41--67\% while improving downstream accuracy. Grid families incorporating both spacing and offset parameters perform comparably, whereas those lacking an offset exhibit clearly higher error, confirming that these two degrees of freedom drive the observed gains. We therefore propose \methodnameshort{} as the standard grid for block-scaled lookup-table formats, replacing fixed grids.
\end{abstract}
]

\printAffiliationsAndNotice{}
\thispagestyle{plain}

\section{Introduction}\label{sec:intro}

\begin{figure*}[t]
\centering
\includegraphics{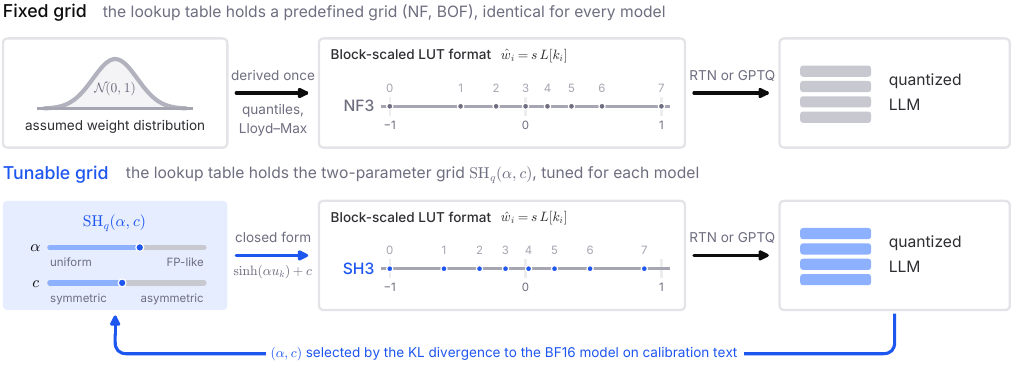}
\caption{\textbf{Fixed grid:} the lookup table holds a grid derived once for an assumed weight distribution (NF, BOF).
\textbf{Tunable grid (this work):} it holds $\methodnameshort(\alpha,c)$, with $(\alpha,c)$ selected for each model by the
calibration KL divergence.  Levels at 3 bits before E4M3 rounding; SH parameters as selected for Qwen3-8B.}
\label{fig:hero}
\end{figure*}

Weight quantization has become a standard way to reduce the memory footprint and serving cost of large language models
(LLMs)~\cite{dettmers2022int8,frantar2023gptq,lin2024awq,dettmers2023qlora}. Specifically, \textit{microscaled}
quantized formats~\cite{rouhani2023microscaling,ocp2023mx,nvidia2025nvfp4} have become very popular: a small block of
weights, typically 16 or 32, shares a scale, and each normalized weight is stored as an index into a \emph{grid} of
$2^q$ values. When that index is decoded through a small lookup table~\cite{killian2026ptq}, the grid need not be an
integer or floating-point grid at all, which raises a basic question: \emph{which} $2^q$ values should the table hold?
Lookup-table formats now range from NF4~\cite{dettmers2023qlora} and BOF4~\cite{blumenberg2025bof4} to the 3-bit
lookup-table instruction of NVIDIA Rubin~\cite{nvidia2026lutb}; while these tables can hold arbitrary values, each
format fixes them in advance.

\paragraph{Optimal grids.} Finding good grids for a given distribution is an old problem. The Lloyd--Max conditions
are necessary conditions for the MSE-optimal scalar quantizer of a known
source~\cite{lloyd1982least,max1960quantizing}. In machine learning, ZipML was among the first to compute
variance-optimal quantization levels directly from the data distribution~\cite{zhang2017zipml}. For LLM weights, this
line of work has produced multiple families of grids: NF4 places its levels at quantiles of a normal
distribution~\cite{dettmers2023qlora}, Student Float derives its levels from a heavier-tailed Student-$t$
distribution~\cite{dotzel2024studentfloat}, BOF4 derives Lloyd-style codebooks that minimize the blockwise
reconstruction error under absmax scaling~\cite{blumenberg2025bof4}, and SqueezeLLM runs sensitivity-weighted
$k$-means for each output channel~\cite{kim2024squeezellm}. Other grids are inherited from number formats, such as INT
and the E2M1 grid of NVFP4~\cite{nvidia2025nvfp4}. A related line of work compares integer and floating-point formats
per layer, or selects between them~\cite{kuzmin2023fp8int8,zhang2023mofq}, and recent formats go further, choosing
between two grids for every block~\cite{cook2026adaptive,egiazarian2026grids}. In most of these formats, the grids are
fixed before the model is seen.

\paragraph{Limits of current approaches.} Fixing the grid in advance has three weaknesses. First, the optimal grid
depends on the bit-width: in our experiments, the best fixed grid is BOFS at 2 bits, NF at 4 and 5 bits, and, in half
of the settings, INT at 6 bits (Section~\ref{sec:results}), so no single fixed grid is a safe choice. Second, the
optimal grid depends on the surrounding format as much as on the weights: the encoding of the block scale in E4M3 and
the search over its values shift the error-minimizing grid (Section~\ref{sec:theory}), whereas NF and BOF are derived
from an assumed weight distribution under plain absmax scaling. Third, and more fundamentally, grids are fit to
mean-squared error, possibly weighted by activation or curvature statistics; however, a lower layerwise error,
weighted or not, does not necessarily yield a lower divergence of the model's
output~\cite{yvinec2023powerquant,blumenberg2025bof4}. Learning the grid for every layer or
channel~\cite{kim2024squeezellm} removes the distributional assumption, but a learned codebook has $2^q$ free levels
per layer and is in practice fit to a layerwise proxy.

\begin{figure*}[t]
\centering
\includegraphics{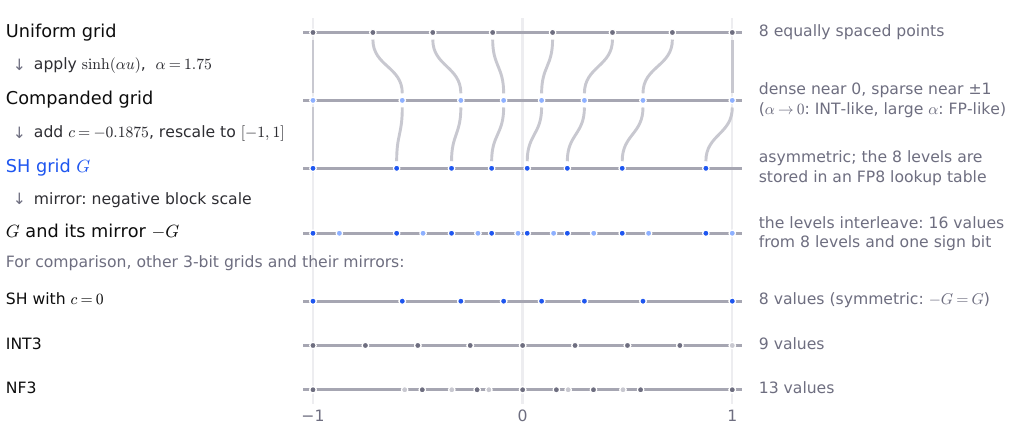}
\caption{Construction of the 3-bit SH grid selected for Qwen3-8B ($\alpha=1.75$, $c=-0.1875$; levels before E4M3
rounding).  A negative block scale selects the mirror image $-G$, whose levels interleave with those of $G$: 16 values
per block, against 8, 9 and 13 for SH with $c=0$, INT3 and NF3.}
\label{fig:sh-construction}
\end{figure*}

\paragraph{This paper.} We take a different approach: rather than fixing the grid in advance or fitting it to a
layerwise proxy, we select it from a two-parameter family against the end-to-end objective. We use a parametrized grid
family, $\methodnameshort(\alpha,c)$, whose $2^q$ levels are obtained by sampling $\sinh(\alpha x)$ on a uniform
lattice and shifting by an offset $c$. Each grid is described by two continuous parameters rather than $2^q$ free
levels, so the space of grids is small enough to be searched by its effect on the model's output
(Figure~\ref{fig:hero}). For each model and bit-width, a scan selects $(\alpha,c)$ by the KL divergence between the
next-token distributions of the quantized and original models on calibration text, and the model is then quantized in
that format with RTN or GPTQ. A fixed 4-bit member of the family, with $\alpha=1.5$ and $c=-0.02$, serves as one of
the read-only codebooks of a methodology for per-layer lookup-table hardware~\cite{killian2026ptq}; here we analyze
the family as a whole, select its parameters per model and width, and evaluate it end to end against fixed grids.
Since only the contents of the lookup table change, $\methodnameshort$ requires no change to storage or decoding, and
we propose it as the default grid of block-scaled lookup-table formats.

\paragraph{The SH grid.} SH is a \emph{companding} quantizer in the classical sense~\cite{gersho1992vector}: uniform
quantization applied after a smooth compressive map (here $\mathrm{asinh}$), so that its levels are dense near zero,
where most normalized weights lie, and sparse in the tails (Figure~\ref{fig:sh-construction}). For $c=0$, the
parameter $\alpha$ sets how strongly it compands, interpolating between an almost uniform, INT-like grid as
$\alpha\to0$ and increasingly geometric, FP-like spacing as $\alpha$ grows and $\sinh$ becomes exponential. The offset
$c$ breaks the odd symmetry of $\sinh$, making the grid asymmetric. Like the grid pairs of recent multi-grid
formats~\cite{cook2026adaptive,egiazarian2026grids}, the otherwise idle sign bit of the E4M3 block scale then selects
between two grids, here the grid and its mirror image. Their levels interleave: for the 3-bit grid selected for
Qwen3-8B, the two provide 16 distinct values, against 8 for SH without an offset, 9 for INT3 and 13 for NF3, at no
storage cost. BOF4-S uses the same mirroring, with the sign fixed by the largest weight of the
block~\cite{blumenberg2025bof4}. All levels are snapped to FP8 (E4M3), so SH is decoded through a $2^q$-entry lookup
table exactly like NF or BOF, and needs no new arithmetic in hardware.

\paragraph{Contributions.}  In summary:
\begin{itemize}
  \item We introduce $\methodnameshort(\alpha,c)$, a two-parameter family of asymmetric companding grids that is
        FP8-representable and lookup-table decodable, and contains near-uniform and near-geometric grids as
        limiting cases. We show that its error-minimizing parameters depend on the encoding and search of the
        block scale as much as on the weights, and that its offset lets the sign of the block scale double the
        number of values available to each block, up to $2^{q+1}$ (Section~\ref{sec:theory}).
  \item We show that these two degrees of freedom, level spacing and offset, account for the gains: two other families
        with both, NormalFloat with a free width and offset and an arctan compander, come within 3\% of SH
        after GPTQ, whereas NormalFloat without an offset, with or without a free width, remains 8.9--16.7\%
        above it, and removing the sign bit, on which the offset relies, raises the KL of SH by 21--38\% at 3
        and 4 bits (Appendices~\ref{app:families} and~\ref{app:flip}). A Lloyd codebook learned for every
        layer matches SH in calibration KL and is 6.5\% lower after GPTQ on Llama-3.1-8B-Instruct, so SH is
        not optimal in an absolute sense; it exposes the two relevant degrees of freedom in closed form.
  \item Across five models from 3B to 35B parameters, including a mixture-of-experts model, at 2 to 6 bits with RTN
        and GPTQ, SH attains the lowest held-out KL divergence in 45 of 50 settings, including all 20 at 3 and
        4 bits, where it is 4.0--34.4\% below the best fixed grid (Table~\ref{tab:main-kl}); differences
        within the measured run-to-run variation of GPTQ are not interpreted (Appendix~\ref{app:xgpu}). The
        expected acceptance rate follows the KL ordering in 47 of the 50 settings, and downstream accuracy
        agrees wherever differences exceed noise, except on the mixture-of-experts model at 2 bits after GPTQ
        (Section~\ref{sec:results}).
  \item At 4.5 bits per weight, the SH4 format lowers the KL divergence of NVFP4 by 40.7--67.4\% and attains higher
        average downstream accuracy in all six model-quantizer combinations evaluated, by 0.1--0.7 points.
        Much of the KL gap comes from the signed, searched block scale; under the same scale, SH4 lowers the
        KL of the best fixed grid by a further 4.6--34.4\% (Table~\ref{tab:nvfp4}). When
        RCO~\cite{helcig2026rco} allocates per-layer bit-widths, SH remains the lowest-KL grid after GPTQ
        (Table~\ref{tab:mixed}); tuning the grid for each layer adds little, so a single grid per model and
        width suffices (Section~\ref{sec:results}).
\end{itemize}
\clearpage
\section{The SH Grid}\label{sec:theory}
\makeatletter
\providecommand{\archp}[2]{\@ifundefined{r@#2}{}{ (#1~\ref{#2})}}
\providecommand{\archs}[2]{\@ifundefined{r@#2}{}{; #1~\ref{#2}}}
\makeatother

$\methodnameshort(\alpha,c)$ (Figure~\ref{fig:sh-construction}) samples $\sinh$ on the uniform $q$-bit lattice
$u_k=-1+2k/(2^q-1)$, adds the offset $c$, and normalizes the largest-magnitude level to 1:
\begin{equation}\label{eq:method-sh}
  y_k=\frac{\sinh(\alpha u_k)+c}{\max_j|\sinh(\alpha u_j)+c|},\quad u_k=-1+\frac{2k}{2^q-1}.
\end{equation}
Each block of 16 weights stores indices into a lookup table holding these levels in E4M3, together with a signed E4M3 scale whose code and sign a per-block search selects (Section~\ref{sec:method-formats}); a negative scale quantizes the block against the mirror image of the grid.  The mirror of the grid is obtained by negating $c$ and is already available through the sign of the block scale, so
$c\le0$.

The grid $\methodnameshort(\alpha,c)$ carries two parameters that act on different aspects of the quantization: the
companding strength $\alpha$ sets the density of levels within the grid, and the offset $c$ breaks the symmetry between
the grid and its reflection.  We analyze each degree of freedom in turn, in the language of classical companding
quantization, and show by measurement that these two degrees of freedom, which $\methodnameshort$ exposes in closed
form, account for its performance: other families with a width and an offset perform comparably, whereas families
without an offset remain clearly behind\archp{Appendix}{app:families}.

\paragraph{Companding.}  Without the offset, the $2^q$ levels of $\methodnameshort(\alpha,0)$ are
\begin{equation}\label{eq:sh-levels}
  y_k=\frac{\sinh(\alpha u_k)}{\sinh\alpha},\qquad u_k=-1+\frac{2k}{2^q-1},
\end{equation}
for $k=0,\dots,2^q-1$, where the points $u_k$ are uniformly spaced in $[-1,1]$.  Equivalently, the levels are uniformly
spaced after the compressor $y\mapsto\mathrm{asinh}(y\sinh\alpha)$, as in a classical companding quantizer.  The
resulting level density,
\begin{equation}\label{eq:sh-density}
  \lambda_\alpha(y)\propto\bigl(1+y^2\sinh^2\alpha\bigr)^{-1/2},
\end{equation}
is nearly constant around the origin, decays as $1/|y|$ towards the ends of the range, and is lower at the edge than at
the center by a factor of $\cosh\alpha$.  The parameter $\alpha$
thus interpolates between uniform quantization ($\alpha\to0$) and geometric, FP-like spacing (large $\alpha$;
Figure~\ref{fig:sh-construction}).

Under the high-resolution approximation, the mean-squared error of a scalar quantizer with level density $\lambda$ on a
source of density $p$ is proportional to $\int p/\lambda^2\,\mathrm{d}y$ and is minimized by
$\lambda\propto p^{1/3}$~\cite{gersho1992vector}.  The SH density has exactly this form for a Student-$t$ source with
two degrees of freedom, so the grid is asymptotically MSE-optimal for that source.  More generally, because
$1/\lambda_\alpha^2$ is quadratic in $y$, the high-resolution error of the family depends on the source only through its
second moment $\sigma^2=\mathrm{E}[y^2]$.  Normalized on $[-1,1]$, the density is
$\lambda_\alpha(y)=\sinh\alpha/\bigl(2\alpha\sqrt{1+y^2\sinh^2\alpha}\bigr)$, so that
$\int p/\lambda_\alpha^2\,\mathrm{d}y=(2\alpha/\sinh\alpha)^2\int p(y)\,(1+y^2\sinh^2\alpha)\,\mathrm{d}y$, and the last
integral equals $1+\sigma^2\sinh^2\alpha$:
\begin{equation}\label{eq:sh-mse}
  \mathrm{MSE}(\alpha)\propto\Bigl(\frac{\alpha}{\sinh\alpha}\Bigr)^{2}\bigl(1+\sigma^2\sinh^2\alpha\bigr).
\end{equation}
After division by the block absmax, the weights of all layers share a single, approximately Gaussian shape; the heavy tails of the raw weights are block-to-block variation of the absmax, which
the per-block scale absorbs.  A single choice of $\alpha$ is therefore appropriate for the entire model by
Eq.~(\ref{eq:sh-mse}).

\paragraph{Dependence on the format.}  The minimizing value of $\alpha$ is determined by the discretizing format as much
as by the weights.  Minimizing the MSE of Eq.~(\ref{eq:sh-mse}) on the empirical block histograms of both 8B models,
with exact scales and unrounded levels, yields a smaller $\alpha$ at 3 bits than the format scan selects.  Under a signed E4M3 block scale pinned to
the block absmax, the per-layer error of Qwen3-8B is instead minimized at larger values of $\alpha$.  When the scale code and sign of every block are free to be chosen, as in the per-block
scale search of Section~\ref{sec:method-formats} but without the channel-norm weights, the minimum lies at $\alpha=1.75$ in 249 of 252 layers
(Figure~\ref{fig:sh-layers}), the value selected by the calibration-KL scan of Section~\ref{sec:method-scans} for this model.
Fixed codebooks such as NF and BOF are designed for other operating points and cannot shift their optimum in this way.

\paragraph{Role of the offset.}  The E4M3 block scale carries a sign bit, so each block is quantized against the grid $G$
or against its reflection $-G$.  For a symmetric grid the two coincide and the sign bit is idle; the offset $c$ breaks
this symmetry.  A block can place its largest-magnitude weight at the longer end of the grid, and at the selected
offset the levels of $-G$ interleave with those of $G$ (Figure~\ref{fig:sh-construction}).  For the 3-bit grid selected for
Qwen3-8B, $G\cup(-G)$ contains 16 distinct values, against 8 for $c=0$, 9 for INT3 and 13 for NF3.  The per-block scale
search selects the orientation minimizing the error weighted by the squared input-channel norms, so the additional
resolution acts primarily on the weights of high-norm channels, which dominate that weighted error.  NF and BOF are asymmetric as
well, but by a fixed amount: NF with a free width remains 13.0--25.9\% above SH in calibration KL, and a free offset closes 74--97\% of
this gap\archp{Table}{tab:families-calib}.  Without the sign, the held-out KL rises in every configuration of Appendix~\ref{app:flip}, by at least 10\%.

\paragraph{Relation to fixed grids.} In terms of Eq.~(\ref{eq:sh-density}), a grid is characterized by its level density $\lambda$ and asymmetry. The baseline grids (Section~\ref{sec:method-grids}) fix both parameters in advance, following one of four approaches:
\begin{itemize}
  \item \textbf{Uniform spacing (INT).} The level density is constant. This configuration is optimal for values distributed uniformly across the range~\cite{gersho1992vector}; however, for bell-shaped weight distributions, it allocates an equal number of levels to sparse tails and dense centers.
  \item \textbf{Geometric spacing (floating point).} Floating-point grids distribute levels geometrically~\cite{micikevicius2022fp8,kuzmin2023fp8int8}, with $\lambda\propto1/|y|$ away from zero, ensuring that the relative rounding error is approximately constant across binades. E2M1 provides fine resolution near zero and coarser resolution in the tails; its symmetry implies that the sign of a scale factor carries no information.
  \item \textbf{Quantiles of a normal distribution (NF).} Each level covers approximately equal probability mass under an assumed normal distribution, yielding $\lambda\propto p$. Consequently, levels are densest near zero and more concentrated there than under the MSE-optimal density $p^{1/3}$. Since the grid includes $0$ and $\pm1$, one side contains one more level than the other.
  \item \textbf{Minimum MSE for Gaussian blocks (BOF, BOFS).} The Lloyd--Max conditions yield the MSE-optimal grid for a specific source, Gaussian blocks normalized by their absmax; its level density approaches $p^{1/3}$ at high resolution. In BOFS, negative levels terminate before $-1$, and the sign of the scale determines the orientation of each block, analogous to SH.
\end{itemize}
SH replaces these fixed choices with two parameters: companding strength $\alpha$~\cite{gersho1992vector} and offset $c$. The parameter $\alpha$ continuously adjusts spacing from uniform (INT, $\alpha\to0$) toward geometric (floating-point grids). Its density, given by Eq.~(\ref{eq:sh-density}), is constant as $\alpha\to0$, approaches $1/|y|$ for large $\alpha$, and corresponds to the MSE-optimal density $p^{1/3}$ of a Student-$t$ source for each $\alpha$. The offset $c$ controls the asymmetry that NF, BOF, and BOFS fix a priori. Both parameters are optimized per model and width using calibration KL divergence (Section~\ref{sec:method-scans}), allowing the density to be determined by the model's output distribution rather than an assumed weight distribution. This enables SH to adapt to each width, where different fixed grids may be optimal (Section~\ref{sec:results}).

\paragraph{Role of the $\sinh$ form.}  Both adaptations analyzed above are available to any smooth two-parameter family;
the measurements below attribute the performance to exactly these two degrees of freedom, which $\methodnameshort$
exposes in closed form.  At 3 bits, SH, the arctan compander, a compander with heavier
tails, and NF with a free width and an offset all lie close to the per-layer Lloyd--Max grid in weight MSE,
whereas NF with a free width but no offset lies clearly above it.  In calibration KL, NF with width and offset is within
0.4--5.0\% of SH and the two other companders within 1.6--6.8\%\archp{Table}{tab:families-calib}; after GPTQ
(single runs), the former and the arctan compander remain within 3.0\%, whereas NF without an offset remains
8.9--16.7\% above SH\archp{Table}{tab:families-gptq}.  A Lloyd codebook learned per layer matches SH in calibration KL
and attains a lower KL after GPTQ on Llama-3.1-8B-Inst., so SH is not optimal in an absolute sense.  Its value is
practical: the levels admit a closed form; its two parameters correspond exactly to the two degrees of freedom measured
above; the uniform and geometric grids arise as limiting cases; the levels are representable in an E4M3 lookup table;
and the two-dimensional parameter space is small enough to be searched directly on the KL divergence of the full model.
\section{Method}\label{sec:method}

Section~\ref{sec:results} reports \emph{configurations}: a model, a grid family, an average code width, a quantizer and
an allocation.  A \emph{format scan} fixes the free parameters of each family once per model and width, on calibration
KL (Section~\ref{sec:method-scans}); the model is then quantized in that format with RTN and with GPTQ, which yields the
\emph{uniform-width} models, or a \emph{mixed} model takes each layer from the uniform-width model of its width
(Section~\ref{sec:method-alloc}).  Each configuration is scored by the expected acceptance rate and the KL divergence
of the quantized model to the BF16 model, both measured on held-out text (Section~\ref{sec:method-eval}).  No choice
behind Tables~\ref{tab:main-ear}, \ref{tab:main-kl}, \ref{tab:mixed} and~\ref{tab:nvfp4} uses that text; earlier
design decisions, such as the adoption of the per-block scale search, rest on experiments evaluated on its first 50K
tokens.

\subsection{Block-scaled lookup-table formats}\label{sec:method-formats}

\paragraph{Storage.}  Each row of a weight matrix is partitioned into blocks of $g=16$ consecutive weights.  A $q$-bit
format stores, per weight $w_i$, an index $k_i$ into a lookup table $L$ of $2^q$ E4M3 values and, per block, a signed
E4M3 scale $s$, so that $\hat w_i=s\,L[k_i]$; the code costs $q+0.5$ bits per weight.  The lookup table (one per format
for the whole model) and a per-layer exponent shift of the scales are not counted in the bit rate.  A grid
$G=\{y_0<\dots<y_{2^q-1}\}$ with $\max_k|y_k|=1$ is a unit codebook: its table stores the levels after a common scaling,
rounded to E4M3.  The \emph{LUT placement factor} $f\in\{1,2^{1/4},2^{1/2},2^{3/4}\}$ sets that scaling; it shifts the
levels against the binades of E4M3 and thereby changes the set of values that survives the rounding, so different
placements realize different effective grids.  In a separate experiment, storing the levels of SH in E4M4 to E4M7 instead of E4M3 changed the KL of both 8B
models by $-1.1\%$ to $+3.6\%$, without a consistent improvement.

\paragraph{Block scale, sign and the lever.}  The naive scale of a block maps its absmax to the largest-magnitude level
of the table.  A negative scale quantizes the block against the reflected grid $-G$ (Section~\ref{sec:theory}).
\emph{The lever} extends this choice to a search: it considers 25
multipliers of the naive scale, geometrically spaced from 0.6 to 1.4, in both signs, rounds each candidate to E4M3, and
keeps the candidate minimizing $\sum_i\mathrm{cn}_i^2(w_i-\hat w_i)^2$, where $\mathrm{cn}_i$ is the RMS of the input
channel of $w_i$ on calibration text.  ScaleSweep~\cite{lin2026scalesweep} searches NVFP4 block scales in a related way.

\subsection{Baselines}\label{sec:method-grids}

We evaluate against baselines defined by existing quantization formats. Each baseline employs a fixed grid determined either by a specific number format or an assumption about weight distribution. With the exception of NVFP4, all baselines utilize the same lookup table, block scale, and lever as SH, ensuring that the grid structure is the sole variable under comparison.

INT$q$~\cite{kuzmin2023fp8int8} employs a two's-complement integer grid $\{-2^{q-1},\dots,2^{q-1}-1\}$ with uniform spacing. We evaluate two variants: one using per-block mirroring under absmax scaling and another under rail scaling. For each configuration, we select the variant yielding lower calibration KL divergence.

E2M1, the element format for NVFP4 and OCP microscaling formats~\cite{ocp2023mx,rouhani2023microscaling}, represents values $\{0,\pm0.5,\pm1,\pm1.5,\pm2,\pm3,\pm4,\pm6\}$. NVFP4~\cite{nvidia2025nvfp4} (4.5 bits per weight) combines E2M1 elements with an unsigned E4M3 scale per 16 weights and an uncounted FP32 tensor scale, consistent with NVIDIA's ModelOpt export; it does not employ a lever.

NF~\cite{dettmers2023qlora} positions grid levels at quantiles of a normal distribution, ensuring that each level captures approximately equal mass from normally distributed weights.

BOF~\cite{blumenberg2025bof4} utilizes the Lloyd--Max codebook~\cite{lloyd1982least,max1960quantizing}, which minimizes mean squared error (MSE). It is computed as the fixed point of Lloyd's iteration on the exact density of normalized weights, where blocks of 16 Gaussian weights are divided by their absmax. BOFS, a signed-absmax variant~\cite{blumenberg2025bof4}, divides each block by its largest-magnitude weight (including sign), mapping this weight to $+1$.

\subsection{Quantizers}\label{sec:method-quant}

All linear layers inside the decoder blocks are quantized, including the projections of the linear-attention blocks and
of the routed experts (one matrix per expert); the embeddings, the output head and the MoE router remain in BF16, and
activations are not quantized.  \emph{RTN} quantizes each layer independently: every block is rounded to its nearest
level under the lever, with channel norms from one BF16 pass over 128 calibration windows.  \emph{GPTQ}
\cite{frantar2023gptq} quantizes the decoder blocks in order, each with Hessians and channel norms from the outputs of
the already quantized preceding blocks; the diagonal of the Hessian is damped by 1\% of its mean, columns are processed
in natural order in lazy batches of 128, and at the first column of each group of 16 the lever refits the scales and
signs of the group's blocks on the error-compensated weights.

\subsection{Format selection on calibration KL}\label{sec:method-scans}

For each model and width, the format scan ranks the candidate formats by the KL divergence of the RTN model to the BF16
model on 64 calibration windows, both distributions coarsened to the 128 tokens most likely under the BF16 model plus
one bucket for the rest.  The scan picks SH's $(\alpha,c)$ by a two-stage box search: a coarse lattice, grown while the
best point lies on its edge, followed by a finer grid around the best point.  $\methodnameshort$ at its scanned
$(\alpha,c)$, BOF, BOFS and NF are then scored at the four placement factors and INT as its two variants, and each
family keeps its best format on these windows; the RTN and GPTQ configurations of a model use exactly these formats
(Appendix~\ref{app:sh-params} lists the scanned SH parameters).  To guard against drift between scan and
configuration, every RTN configuration re-scores its calibration KL and is refused if the value differs from the
scan's by more than 1\% or $2\cdot10^{-4}$ nats, whichever is larger.

\subsection{Per-layer bit-width allocation}\label{sec:method-alloc}

The uniform-width models of the two 8B models, at $q=2,\dots,6$, store the quantized weights of every layer, for RTN
and GPTQ alike.  The GPTQ weights come from runs on the RTX A6000 with the formats and calibration windows of
Table~\ref{tab:main-kl}: for Qwen3-8B the very runs of that table, and for Llama-3.1-8B-Instruct a second run.  A
mixed model reuses these stored layers: each layer takes the width of its uniform-width model, and no quantization is
rerun, so a uniform-width model is exactly the assignment $q_l=q$ for all $l$.  The target $B$ fixes the average code
width $\sum_l n_l q_l/\sum_l n_l$, with $n_l$ the number of weights of layer $l$.  RCO~\cite{helcig2026rco} optimizes
a categorical distribution over the five bit-widths of each layer, constrained to assignments whose expected average
width is $B$: for each Gumbel sample it solves the multiple-choice knapsack that projects the sample onto this
constraint, and back-propagates the KL divergence of the resulting assignment to the unquantized model on calibration
text (600 steps, 4 Gumbel samples, one seed).  Table~\ref{tab:mixed} reports $B=3$, 4 and 5.

\subsection{Evaluation}\label{sec:method-eval}

\paragraph{Data.}  All text is FineWeb-Edu (sample-10BT), split by document: evaluation $[0,5000)$, validation
$[50000,55000)$ (used only to tune RCO) and calibration from document 100000 on.  Each split is tokenized as one stream
and cut into consecutive 2048-token windows.

\paragraph{Held-out KL.}  Every KL in Tables~\ref{tab:main-kl}, \ref{tab:mixed}, \ref{tab:nvfp4}
and~\ref{tab:xgpu} is
\begin{equation}\label{eq:method-kl}
  \mathrm{KL}=\frac{1000}{N}\sum_{t=1}^{N}\sum_{v}p_t(v)\ln\frac{p_t(v)}{\hat p_t(v)}\quad\mathrm{(mn)},
\end{equation}
the full-vocabulary divergence of the next-token distribution $\hat p$ of the quantized model from that $p$ of the BF16
model, in millinats, over the $N$ next-token positions of the first 500K evaluation tokens (245 windows).  Both models
compute the output head and the softmax in FP32.  The Llama models get a BOS token prepended to every window; the
Qwen tokenizers have no BOS.

\paragraph{Expected acceptance rate.}  Table~\ref{tab:main-ear} reports, on the same positions and FP32 logits, the
top-10 expected acceptance rate~\cite{helcig2026slq}
\begin{equation}\label{eq:method-ear}
  \mathrm{EAR}=\frac{100}{N}\sum_{t=1}^{N}\sum_{v\in T_t}\min\bigl(\bar p_t(v),\bar{\hat p}_t(v)\bigr)\quad(\%),
\end{equation}
where $T_t$ holds the ten tokens most likely under the BF16 model and $\bar p_t$ and $\bar{\hat p}_t$ are the two
distributions renormalized over $T_t$.  In speculative decoding it is the acceptance probability of a token proposed
by the BF16 model, restricted to these ten tokens; higher is better, and 100\% corresponds to identical
distributions.

\paragraph{Statistics and hardware.}  Every configuration is a single run.  Two configurations are compared by their
relative difference in KL, or, for EAR, by their difference in percentage points; the 95\% interval of a difference
comes from 10{,}000 bootstrap resamples of the 245 windows, and the interval of a difference between two such
differences from the same resamples.  A difference whose 95\% interval includes zero is not interpreted.
Reruns with the same code, inputs and GPU type are bit-identical, and scoring the same weights on a different GPU
type changes the KL only little.  GPTQ is an exception: it is sensitive to small numerical differences, and on a
different GPU type, or on the same GPU with a different lazy-batch width (which only reorders floating-point sums),
it returns a different quantization (Appendix~\ref{app:xgpu}).  Differences in KL below about 3\% between single
configurations are therefore not interpreted, and neither are differences in EAR within the spread of two such GPTQ
runs: 0.19 percentage points at 2 and 3 bits and 0.02 at 4 to 6 bits on Llama-3.1-8B-Instruct.  Except for
Llama-3.2-3B-Instruct, which ran entirely on RTX 3090, the format scans ran on H200 (Qwen3-8B, Qwen3.8-27B), L40S (Llama-3.1-8B-Instruct) and RTX A6000 (Qwen3.6-35B-A3B, except at 2, 5 and 6
bits, which ran on RTX 3090); the GPTQ configurations on RTX A6000 (Qwen3-8B, Qwen3.6-35B-A3B) and L40S
(Llama-3.1-8B-Instruct, Qwen3.8-27B); and the RTN configurations, the layer databases of the mixed models and RCO on
RTX A6000.
\begingroup
\makeatletter
\let\origtable\table \let\endorigtable\endtable
\renewenvironment{table}[1][]{\origtable[!htb]}{\endorigtable}
\makeatother
\section{Results}\label{sec:results}

\begin{table*}[t]
\centering
\caption{KL divergence (mn, lower is better) to the BF16 model on 500K held-out tokens, one format for all layers.
Each width adds 0.5 bpw for the E4M3 block scale.  Bold: best per model and row.}
\label{tab:main-kl}
{\setlength{\tabcolsep}{3pt}%
\resizebox{\textwidth}{!}{%
\begin{tabular}{@{}llrrrrrrrrrrrrrrrrrrrr@{}}
\toprule
 & & \multicolumn{4}{c}{Llama-3.2-3B-Inst.} & \multicolumn{4}{c}{Qwen3-8B} & \multicolumn{4}{c}{Llama-3.1-8B-Inst.} & \multicolumn{4}{c}{Qwen3.8-27B} & \multicolumn{4}{c}{Qwen3.6-35B-A3B} \\
\cmidrule(lr){3-6}\cmidrule(lr){7-10}\cmidrule(lr){11-14}\cmidrule(lr){15-18}\cmidrule(lr){19-22}
 & & SH & BOFS & NF & INT & SH & BOFS & NF & INT & SH & BOFS & NF & INT & SH & BOFS & NF & INT & SH & BOFS & NF & INT \\
\midrule
\multicolumn{22}{@{}l}{\textbf{2 bits} \footnotesize(2.5 bpw incl.\ scale)} \\
 uniform & RTN & \textbf{744.1} & 939.8 & 1228.2 & 1209.7 & \textbf{520.6} & 561.8 & 601.8 & 618.2 & \textbf{643.2} & 819.0 & 1001.7 & 955.4 & \textbf{212.2} & 242.2 & 286.0 & 273.8 & \textbf{250.6} & 289.7 & 336.7 & 328.7 \\
 uniform & GPTQ & \textbf{577.0} & 782.7 & 813.6 & 985.9 & 379.8 & \textbf{378.2} & 432.5 & 405.3 & \textbf{509.0} & 628.8 & 727.8 & 685.7 & \textbf{194.5} & 212.5 & 249.6 & 230.1 & \textbf{208.2} & 224.4 & 263.6 & 240.5 \\
\midrule
\multicolumn{22}{@{}l}{\textbf{3 bits} \footnotesize(3.5 bpw incl.\ scale)} \\
 uniform & RTN & \textbf{94.45} & 122.7 & 112.2 & 168.1 & \textbf{78.31} & 87.23 & 88.60 & 109.6 & \textbf{90.82} & 107.0 & 105.2 & 138.0 & \textbf{44.44} & 48.28 & 49.83 & 58.06 & \textbf{55.40} & 63.74 & 62.45 & 77.77 \\
 uniform & GPTQ & \textbf{82.66} & 116.8 & 99.31 & 159.3 & \textbf{61.90} & 66.94 & 67.60 & 81.51 & \textbf{81.46} & 98.25 & 95.94 & 118.8 & \textbf{38.45} & 40.07 & 41.40 & 46.72 & \textbf{46.92} & 49.45 & 49.63 & 57.83 \\
\midrule
\multicolumn{22}{@{}l}{\textbf{4 bits} \footnotesize(4.5 bpw incl.\ scale)} \\
 uniform & RTN & \textbf{20.36} & 27.86 & 25.60 & 35.56 & \textbf{18.35} & 22.82 & 21.61 & 28.89 & \textbf{21.21} & 26.09 & 25.47 & 34.27 & \textbf{10.99} & 12.42 & 12.03 & 15.24 & \textbf{16.68} & 19.67 & 18.48 & 24.26 \\
 uniform & GPTQ & \textbf{16.98} & 30.29 & 25.90 & 37.48 & \textbf{14.46} & 16.14 & 15.78 & 19.94 & \textbf{19.81} & 25.13 & 25.63 & 32.93 & \textbf{9.33} & 9.81 & 10.11 & 11.74 & \textbf{14.44} & 15.34 & 15.13 & 18.15 \\
\midrule
\multicolumn{22}{@{}l}{\textbf{5 bits} \footnotesize(5.5 bpw incl.\ scale)} \\
 uniform & RTN & \textbf{5.68} & 7.16 & 6.67 & 9.91 & \textbf{5.36} & 6.55 & 6.08 & 8.35 & \textbf{5.92} & 6.87 & 6.70 & 8.96 & \textbf{3.19} & 3.51 & 3.43 & 4.27 & \textbf{7.28} & 7.89 & 7.48 & 9.17 \\
 uniform & GPTQ & \textbf{4.62} & 7.85 & 6.57 & 11.49 & \textbf{4.22} & 4.73 & 4.52 & 5.69 & \textbf{5.26} & 7.65 & 6.74 & 9.81 & \textbf{2.71} & 2.78 & 2.88 & 3.31 & 6.48 & 6.58 & \textbf{6.42} & 7.32 \\
\midrule
\multicolumn{22}{@{}l}{\textbf{6 bits} \footnotesize(6.5 bpw incl.\ scale)} \\
 uniform & RTN & \textbf{2.17} & 2.74 & 2.46 & 2.59 & \textbf{2.17} & 2.75 & 2.46 & 2.54 & \textbf{2.27} & 2.80 & 2.54 & 2.58 & 1.36 & 1.71 & 1.47 & \textbf{1.32} & \textbf{4.48} & 4.82 & 4.71 & 4.61 \\
 uniform & GPTQ & \textbf{1.76} & 2.81 & 2.28 & 2.93 & \textbf{1.73} & 2.06 & 1.91 & 1.79 & \textbf{1.99} & 2.62 & 2.34 & 3.33 & 1.16 & 1.33 & 1.18 & \textbf{1.09} & 4.11 & 4.36 & 4.19 & \textbf{4.06} \\
\bottomrule
\end{tabular}}}
\end{table*}

\begin{figure*}[t]
\centering
\includegraphics{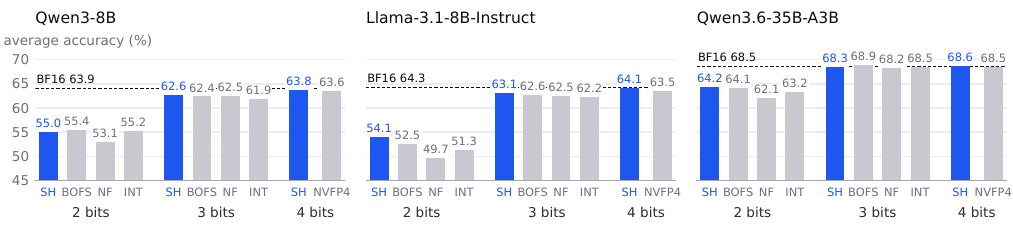}
\caption{Average accuracy (\%) over six downstream benchmarks after RTN with one format for all layers; dashed line: BF16 model.  The tasks, per-task accuracies and the GPTQ results are in Appendix~\ref{app:bench}.}
\label{fig:bench-avg}
\end{figure*}

We evaluate SH against the fixed grids NF, BOFS, and INT on Llama-3.2-3B-Instruct, Qwen3-8B, Llama-3.1-8B-Instruct, Qwen3.8-27B, and the mixture-of-experts model Qwen3.6-35B-A3B. Experiments are conducted at code widths of 2 to 6 bits, applying a single format to all linear layers. Each model is quantized using both RTN and GPTQ (Section~\ref{sec:method-quant}). All formats incorporate an additional 0.5 bpw for the E4M3 block scale, ensuring that grids of identical width have equal size. Table~\ref{tab:main-kl} reports held-out KL divergence relative to the BF16 model; Figure~\ref{fig:bench-avg} presents accuracy on six downstream benchmarks; Table~\ref{tab:mixed} details KL divergence when RCO optimizes per-layer widths; and Table~\ref{tab:nvfp4} compares performance with NVFP4 at 4.5 bpw. All results are from single runs. We do not interpret KL differences below 3\%, a magnitude attributable to GPTQ run-to-run variation (Appendix~\ref{app:xgpu}), or differences whose 95\% bootstrap confidence interval includes zero (Section~\ref{sec:method-eval}).

\paragraph{Held-out KL.}
Table~\ref{tab:main-kl} reveals three primary patterns. First, SH achieves the lowest KL in 45 of the 50 settings where all four grids are available, including all 20 settings at 3 and 4 bits, where it outperforms the best fixed grid by 4.0--34.4\%. The average margin over the best fixed grid is 14\%, 11\%, and 14\% at 2, 3, and 4 bits, respectively, decreasing to 11\% and 7\% at 5 and 6 bits. Three of the five exceptions fall within GPTQ run-to-run variation: BOFS on Qwen3-8B at 2 bits (0.4\% below SH), and NF and INT on Qwen3.6-35B-A3B at 5 and 6 bits (0.9\% and 1.2\%, respectively). The other two are INT at 6 bits on Qwen3.8-27B, 3.3\% below SH under RTN and 5.6\% below SH under GPTQ. Second, no single fixed grid dominates across all widths: BOFS is the best fixed grid in all ten settings at 2 bits; NF leads in 17 of 20 settings at 4 and 5 bits; and INT leads in five of ten settings at 6 bits. The parameters selected for SH vary across models and widths (Appendix~\ref{app:sh-params}). Third, the advantage of SH persists under both quantizers. GPTQ reduces the KL of SH by 7--27\% relative to RTN, maintaining SH as the lowest-KL method in 21 of 25 GPTQ settings, compared to 24 of 25 under RTN. GPTQ narrows the margin for Qwen3-8B, Qwen3.8-27B, and Qwen3.6-35B-A3B (to at most 8.3\%, 8.5\%, and 7.2\%, respectively) but widens it for both Llama models, from 11.8--20.8\% under RTN to 16.8--34.4\% on Llama-3.2-3B-Instruct and from 10.5--21.5\% to 15.0--21.9\% on Llama-3.1-8B-Instruct. The expected acceptance rate (EAR; Appendix~\ref{app:ear}) matches the KL ordering in 47 of the 50 settings, and SH exhibits the highest EAR in 46 of them.

\paragraph{Downstream accuracy.}
Figure~\ref{fig:bench-avg} displays average accuracy for RTN models across six benchmarks; Appendix~\ref{app:bench} lists individual tasks and GPTQ results. Under both quantizers, all grids remain within 2.1 points of BF16 at 3 bits and within 0.9 points at 4 bits. At 4.5 bpw, SH4 outperforms NVFP4 on both 8B models by 0.4 and 0.5 points after GPTQ, and by 0.2 and 0.7 points under RTN, while remaining within 0.3 points of BF16; the difference is significant for Llama-3.1-8B-Instruct. On Qwen3.6-35B-A3B, SH4 exceeds NVFP4 by 0.1 points under RTN and 0.4 points after GPTQ, within noise, and stays within 0.1 points of BF16. At 3 bits, SH achieves the highest average accuracy on both 8B models under both quantizers, exceeding the next-best grid by 0.1--0.4 points, though no such gap is statistically significant. On Qwen3.6-35B-A3B, BOFS exceeds SH by 0.4 points (0.5 under RTN), a difference within noise. At 2 bits, the grids diverge. On Llama-3.1-8B-Instruct, SH leads BOFS by 1.5 points after GPTQ and 1.6 under RTN, and NF/INT by 2.8--4.3 points; all differences are significant. On Qwen3-8B, BOFS exceeds SH by 0.6 points (0.5 under RTN), within noise. On Qwen3.6-35B-A3B, BOFS and INT exceed SH by 1.3 and 1.2 points after GPTQ, respectively, a significant difference despite SH having lower KL (208.2 vs. 224.4 and 240.5 mn); under RTN, SH and BOFS are comparable (64.2 vs. 64.1). Where differences exceed noise, SH's lower KL generally translates to higher accuracy, with the exception of 2-bit GPTQ models for the mixture-of-experts model.

\begin{table}[t]
\centering
\caption{Held-out KL (mn) of the 8B models with one code width for all layers (\emph{uniform}) and with per-layer
widths chosen by RCO at the same average width (\emph{mixed}, one seed), on RTX A6000.  Bold: best per model and row.}
\label{tab:mixed}
{\setlength{\tabcolsep}{3pt}%
\resizebox{\columnwidth}{!}{%
\begin{tabular}{@{}llrrrrrrrr@{}}
\toprule
 & & \multicolumn{4}{c}{Qwen3-8B} & \multicolumn{4}{c}{Llama-3.1-8B-Inst.} \\
\cmidrule(lr){3-6}\cmidrule(lr){7-10}
 & & SH & BOFS & NF & INT & SH & BOFS & NF & INT \\
\midrule
\multicolumn{10}{@{}l}{\textbf{3 bits}} \\
 RTN & uniform & \textbf{78.31} & 87.23 & 88.60 & 109.6 & \textbf{90.82} & 107.0 & 105.2 & 138.0 \\
 RTN & mixed & \textbf{77.79} & 80.26 & 83.84 & 101.1 & \textbf{85.12} & 91.09 & 96.23 & 107.1 \\
 GPTQ & uniform & \textbf{61.90} & 66.94 & 67.60 & 81.51 & \textbf{83.66} & 98.18 & 94.57 & 119.4 \\
 GPTQ & mixed & \textbf{60.49} & 62.77 & 64.25 & 74.46 & \textbf{78.18} & 78.39 & 85.62 & 89.53 \\
\midrule
\multicolumn{10}{@{}l}{\textbf{4 bits}} \\
 RTN & uniform & \textbf{18.35} & 22.82 & 21.61 & 28.89 & \textbf{21.21} & 26.09 & 25.47 & 34.27 \\
 RTN & mixed & 20.36 & \textbf{19.29} & 28.83 & 24.61 & \textbf{19.40} & 20.49 & 21.10 & 24.98 \\
 GPTQ & uniform & \textbf{14.46} & 16.14 & 15.78 & 19.94 & \textbf{19.88} & 25.39 & 25.62 & 33.30 \\
 GPTQ & mixed & \textbf{13.55} & 14.56 & 14.37 & 17.65 & \textbf{16.66} & 18.33 & 18.31 & 22.96 \\
\midrule
\multicolumn{10}{@{}l}{\textbf{5 bits}} \\
 RTN & uniform & \textbf{5.36} & 6.55 & 6.08 & 8.35 & \textbf{5.92} & 6.87 & 6.70 & 8.96 \\
 RTN & mixed & \textbf{4.86} & 5.62 & 5.37 & 6.62 & \textbf{5.24} & 5.91 & 5.87 & 7.07 \\
 GPTQ & uniform & \textbf{4.22} & 4.73 & 4.52 & 5.69 & \textbf{5.25} & 7.77 & 6.76 & 9.78 \\
 GPTQ & mixed & \textbf{3.86} & 4.15 & 4.00 & 4.74 & \textbf{4.69} & 5.41 & 5.14 & 7.08 \\
\bottomrule
\end{tabular}}}
\end{table}
\paragraph{Per-layer bit widths.}
In Table~\ref{tab:mixed}, RCO optimizes per-layer widths for the two 8B models at average widths of 3, 4, and 5 bits (Section~\ref{sec:method-alloc}). After GPTQ, every mixed model exhibits lower KL than its uniform-width counterpart by 2.3--31.1\%. Fixed grids benefit more from this allocation than SH; in each of the six model-width combinations, the reduction is smallest for SH (2.3--16.2\%). Consequently, the gap between grids narrows from 6.6--22.4\% to 0.3--9.0\%. SH remains the lowest-KL mixed model in all six combinations, with its smallest margin, 0.3\% over BOFS on Llama-3.1-8B-Instruct at 3 bits, within noise. This suggests that superior grids leave less residual error for allocation to mitigate. Under RTN, this pattern holds in five of six combinations, where SH is 3.1--10.8\% below the best fixed grid. On Qwen3-8B at 4 bits, mixed SH and NF models perform worse than their uniform-width counterparts by 10.9\% (driven by a single evaluation window) and 33.4\%, respectively, with BOFS achieving the lowest KL.

\begin{table}[t]
\centering
\caption{4-bit formats vs NVFP4, all at 4.5 bpw.  NVFP4: E2M1 codes, unsigned E4M3 block scale without scale
search, FP32 tensor scale (not counted).  Bold: best per row.}
\label{tab:nvfp4}
\resizebox{\columnwidth}{!}{%
\begin{tabular}{@{}llrrrrr@{}}
\toprule
 & & NVFP4 & SH & BOFS & NF & INT \\
\midrule
Llama-3.2-3B-Inst. & RTN & 52.18 & \textbf{20.36} & 27.86 & 25.60 & 35.56 \\
Llama-3.2-3B-Inst. & GPTQ & 52.04 & \textbf{16.98} & 30.29 & 25.90 & 37.48 \\
Qwen3-8B & RTN & 48.70 & \textbf{18.35} & 22.82 & 21.61 & 28.89 \\
Qwen3-8B & GPTQ & 32.65 & \textbf{14.46} & 16.14 & 15.78 & 19.94 \\
Llama-3.1-8B-Inst. & RTN & 52.26 & \textbf{21.21} & 26.09 & 25.47 & 34.27 \\
Llama-3.1-8B-Inst. & GPTQ & 52.91 & \textbf{19.81} & 25.13 & 25.63 & 32.93 \\
Qwen3.8-27B & RTN & 26.88 & \textbf{10.99} & 12.42 & 12.03 & 15.24 \\
Qwen3.8-27B & GPTQ & 18.24 & \textbf{9.33} & 9.81 & 10.11 & 11.74 \\
Qwen3.6-35B-A3B & RTN & 36.64 & \textbf{16.68} & 19.67 & 18.48 & 24.26 \\
Qwen3.6-35B-A3B & GPTQ & 24.35 & \textbf{14.44} & 15.34 & 15.13 & 18.15 \\
\bottomrule
\end{tabular}}
\end{table}
\paragraph{Comparison with NVFP4.}
At 4.5 bpw, SH4 achieves 40.7--67.4\% lower KL than NVFP4 across all ten settings in Table~\ref{tab:nvfp4}. A substantial portion of this gap stems from block scaling: NVFP4 rounds the absolute maximum scale to an unsigned E4M3 value, whereas lookup-table formats search for signed scales using the lever (Section~\ref{sec:method-formats}). With the lever, even INT4 outperforms NVFP4 by 25.4--43.3\%. Among formats sharing the lever, the best fixed grid is 37.9--55.6\% below NVFP4, and SH4 further reduces KL by 4.6--34.4\% relative to this best fixed grid.

\paragraph{Sign of the block scale.}
Without the sign bit of the block scale, the held-out KL of every grid rises on both 8B models (Appendix~\ref{app:flip}).  At 3 and 4 bits, SH, whose offset relies on the sign, loses more than NF, BOF and INT in seven of the eight settings (21--38\% against 10--31\%), but it remains below the best of them in all eight settings.

\paragraph{Per-layer grids.}
Tuning per-layer grids around global choices rarely reduces KL after GPTQ, so using a single grid per model and width incurs minimal loss (Appendix~\ref{app:per-layer}).
\endgroup
\FloatBarrier
\section{Related Work}\label{sec:related}

\paragraph{Grid design.}  Choosing the reconstruction levels of a scalar quantizer is a classical problem.  The
Lloyd--Max conditions are necessary for the MSE-optimal scalar quantizer of a known source~\cite{lloyd1982least,max1960quantizing}, and
Bennett's observation is that a nonuniform quantizer is equivalent to uniform quantization between a compressor and
an expander, with the high-resolution distortion determined by the point density of the levels~\cite{bennett1948spectra,gray1998quantization}.
For neural networks, learnable piecewise-linear companding grids were developed for convolutional
networks~\cite{yamamoto2021lcq}.  For LLM weights, NF4 places its levels at normal
quantiles~\cite{dettmers2023qlora}, Student Float fits a Student-$t$ distribution~\cite{dotzel2024studentfloat}, and BOF4 applies Lloyd iterations to block-normalized weights, with the error measured on the original weights~\cite{blumenberg2025bof4}.  Other methods learn a codebook for every output channel
by sensitivity-weighted $k$-means~\cite{kim2024squeezellm}.
Most of these grids are fit to weight-space or layerwise errors, and a lower
reconstruction error does not necessarily yield a better model~\cite{yvinec2023powerquant,blumenberg2025bof4}.

\paragraph{Parametric grid families.}  The work closest to \methodnameshort{} consists of grid families whose levels are generated
by a few continuous parameters.  PowerQuant maps a uniform lattice through a power function whose exponent
is chosen by the weight error~\cite{yvinec2023powerquant};
M-ANT selects, for every group of 128 weights, a member of a one-parameter family spanning power-of-two,
floating-point, and integer grids~\cite{hu2025mant}; and the concurrent CubicQuant fits a two-parameter cubic
level function to every group of 128 to 512 weights by weight MSE, evaluated on synthetic data~\cite{gao2026cubicquant}.
Orr et al.\ construct grids whose point density is proportional to $p^{1/3}$ for normal, Laplace, and Student-$t$
weights~\cite{orr2025optimal}; in their parametrization, \methodnameshort{} corresponds to the limiting Student-$t$ source with two
degrees of freedom, admissible only on the bounded range produced by block scaling.  These families are
symmetric, or their parameters are derived from weight statistics and layerwise errors.  To our knowledge, \methodnameshort{} is the
first parametric grid whose parameters are selected by the KL divergence of the full model.

\paragraph{Block-scaled formats and per-block grid selection.}  NVFP4 pairs E2M1 elements with an E4M3 scale for
every 16 elements~\cite{nvidia2025nvfp4}.  Several methods improve the scales of a fixed E2M1 grid, by mapping the
block maximum to a smaller level~\cite{cook2025fouroversix}, or by searching E4M3 scale codes under a plain or
activation-weighted error~\cite{egiazarian2026bridging,lin2026scalesweep}; our per-block scale search belongs to this group.  Other formats
let each block or group choose among several grids: LO-BCQ stores an explicit
selector among learned codebooks~\cite{elangovan2025lobcq}.  Because NVFP4 leaves the
sign bit of its E4M3 scale unused, recent formats spend this bit on a selector: IF4 and MixFP4 choose between FP4
and INT4~\cite{cook2026adaptive,zou2026mixfp4}, RaZeR remaps the redundant zero to a special value~\cite{chen2026razer},
AAAC chooses between two learned tables~\cite{islambouli2026aaac}, and Grid Games analyzes such grid pairs and shows
that their advantage vanishes as the block size grows~\cite{egiazarian2026grids}.  Signed scales have also been used
directly: BOF4-S fixes the sign from the largest weight of the block~\cite{blumenberg2025bof4}, signmax scaling
stores the signed block maximum~\cite{orr2025optimal}, and signed symmetric quantization selects the sign for
integer grids~\cite{colbert2026signed}.  In \methodnameshort{} the sign bit retains its meaning: with $c\neq0$, a single lookup table
and its reflection provide up to $2^{q+1}$ distinct values, and the sign is selected jointly with the scale code.

\paragraph{Bit allocation.}  Mixed-precision quantization assigns each layer a bit-width, chosen by a linear model of
the loss in the layerwise errors~\cite{malinovskii2025higgs} or by Shapley-value estimates of layer
interactions~\cite{zhao2025coopq,helcig2026slq}.  Such proxy models are accurate near full precision but degrade
at low bit-widths, where interactions between layers become significant~\cite{sieberling2025evopress}.
Methods that optimize the end-to-end divergence directly use evolutionary search~\cite{sieberling2025evopress}
or gradient-based search under an exact budget constraint~\cite{helcig2026rco}.
A few allocators also choose the format: FLIQS between integer and floating-point formats during
quantization-aware training~\cite{dotzel2024fliqs}, and Q-Palette
among quantizer types with fractional bit-widths~\cite{lee2025qpalette}.
We use RCO to allocate per-layer bit-widths for each grid (Section~\ref{sec:method-alloc}).

\paragraph{Rounding and vector quantization.}  Once the grid is fixed, the rounding of the weights can
itself be optimized.  GPTQ~\cite{frantar2023gptq}
compensates the error of each quantized weight with second-order updates,
while YAQA targets the end-to-end loss~\cite{tseng2026yaqa}.  These
methods are largely orthogonal to the grid: GPTQ requires only a nearest-level rounding step and therefore applies
to any lookup-table grid, which is how we use it.  Vector and trellis quantizers~\cite{egiazarian2024aqlm,tseng2024qtip}
attain the highest accuracy at 2 to 3 bits per weight, at a higher
decoding cost than one table lookup per weight; scalar lookup-table formats target hardware that decodes every
weight independently.  Learned discrete codes~\cite{dadgarnia2026gsq} keep one table lookup per weight and optimize which level each weight takes, together with the scales (Appendix~\ref{app:gsq}).

\paragraph{Lookup-table decoding and evaluation.}  On the decoding side, kernels that materialize the codebook as
a table of weight values and apply a standard matrix multiplication~\cite{guo2024flute} execute arbitrary grids at
a speed close to that of optimized integer kernels for small batch sizes.  The tensor-core instruction added by
NVIDIA's Rubin architecture decodes 3-bit weight indices through an eight-entry E4M3 table~\cite{nvidia2026lutb},
the table type of the 3-bit formats studied here; its block scale is absent or an unsigned UE8M0 value per 32 weights, whereas our formats use a signed E4M3 scale per 16 weights.  On the evaluation side, benchmark accuracy can
remain unchanged while the answers a model produces change, and the KL divergence to the unquantized model tracks
such changes~\cite{dutta2024accuracy}.  We therefore report the KL divergence throughout; close to the lossless
regime it is, however, a coarse criterion for ranking models that are near-identical~\cite{nikolic2026displacement}.
\section{Conclusion}\label{sec:conclusion}

Block-scaled lookup-table formats typically leave the quantization grid unconstrained; however, existing implementations fix this grid in advance, and the optimal fixed grid varies with bit width. \methodnameshort{} replaces this ad hoc selection with a closed-form family parameterized by two variables: a companding strength $\alpha$, which interpolates level spacing between uniform and floating-point-like distributions, and an offset $c$, which allows the sign bit of the E4M3 block scale to select between the grid and its mirror image. By modifying only the lookup table, \methodnameshort{} achieves the lowest KL divergence across all settings at 3 and 4 bits, and in most settings at 2, 5, and 6 bits. This performance holds across five models ranging from 3B to 35B parameters, and \methodnameshort{} persists as the lowest-KL grid under mixed precision following GPTQ quantization. At a precision of 4.5 bits per weight, the \methodnameshort{} format reduces the KL divergence of NVFP4 by 41--67\% and improves average downstream accuracy on both 8B models. Empirical results indicate that families incorporating both spacing and offset parameters perform comparably, whereas those lacking an offset remain clearly inferior; thus, these two degrees of freedom account for the observed gains. Given that no single fixed grid is optimal across all bit widths and \methodnameshort{} requires no changes to storage or decoding logic, we propose it as the default grid for block-scaled lookup-table formats.

\bibliographystyle{mlsys2027}
\bibliography{references}

\appendix
\newpage

\section{Expected Acceptance Rate}\label{app:ear}

\begin{table*}[t]
\centering
\caption{Expected acceptance rate (EAR, \%, higher is better) to the BF16 model on 500K held-out tokens, one format for
all layers.  Each width adds 0.5 bpw for the E4M3 block scale.  Bold: best per model and row.}
\label{tab:main-ear}
{\setlength{\tabcolsep}{3pt}%
\resizebox{\textwidth}{!}{%
\begin{tabular}{@{}llrrrrrrrrrrrrrrrrrrrr@{}}
\toprule
 & & \multicolumn{4}{c}{Llama-3.2-3B-Inst.} & \multicolumn{4}{c}{Qwen3-8B} & \multicolumn{4}{c}{Llama-3.1-8B-Inst.} & \multicolumn{4}{c}{Qwen3.8-27B} & \multicolumn{4}{c}{Qwen3.6-35B-A3B} \\
\cmidrule(lr){3-6}\cmidrule(lr){7-10}\cmidrule(lr){11-14}\cmidrule(lr){15-18}\cmidrule(lr){19-22}
 & & SH & BOFS & NF & INT & SH & BOFS & NF & INT & SH & BOFS & NF & INT & SH & BOFS & NF & INT & SH & BOFS & NF & INT \\
\midrule
\multicolumn{22}{@{}l}{\textbf{2 bits} \footnotesize(2.5 bpw incl.\ scale)} \\
 uniform & RTN & \textbf{70.44} & 68.00 & 63.92 & 64.03 & \textbf{75.48} & 74.70 & 73.92 & 73.19 & \textbf{73.39} & 70.56 & 67.47 & 67.88 & \textbf{84.18} & 83.08 & 81.62 & 81.68 & \textbf{82.78} & 81.51 & 80.05 & 80.19 \\
 uniform & GPTQ & \textbf{73.96} & 71.27 & 70.47 & 67.76 & \textbf{79.22} & 79.16 & 77.70 & 78.30 & \textbf{76.56} & 74.38 & 72.54 & 73.28 & \textbf{85.09} & 84.29 & 83.01 & 83.62 & \textbf{84.46} & 83.76 & 82.44 & 83.12 \\
\midrule
\multicolumn{22}{@{}l}{\textbf{3 bits} \footnotesize(3.5 bpw incl.\ scale)} \\
 uniform & RTN & \textbf{88.85} & 87.41 & 87.92 & 85.24 & \textbf{90.69} & 90.00 & 90.06 & 88.70 & \textbf{89.92} & 88.95 & 89.12 & 87.35 & \textbf{93.05} & 92.73 & 92.62 & 91.93 & \textbf{92.01} & 91.35 & 91.51 & 90.43 \\
 uniform & GPTQ & \textbf{89.94} & 88.00 & 88.98 & 86.02 & \textbf{91.97} & 91.71 & 91.61 & 90.79 & \textbf{90.94} & 90.00 & 90.15 & 88.85 & \textbf{93.82} & 93.64 & 93.55 & 93.10 & \textbf{92.84} & 92.61 & 92.61 & 91.99 \\
\midrule
\multicolumn{22}{@{}l}{\textbf{4 bits} \footnotesize(4.5 bpw incl.\ scale)} \\
 uniform & RTN & \textbf{94.84} & 94.01 & 94.26 & 93.21 & \textbf{95.52} & 94.95 & 95.11 & 94.29 & \textbf{95.21} & 94.65 & 94.73 & 93.86 & \textbf{96.64} & 96.41 & 96.46 & 95.98 & \textbf{95.68} & 95.31 & 95.44 & 94.75 \\
 uniform & GPTQ & \textbf{95.46} & 93.93 & 94.44 & 93.24 & \textbf{96.19} & 95.97 & 96.00 & 95.49 & \textbf{95.65} & 95.05 & 95.00 & 94.32 & \textbf{97.04} & 96.96 & 96.92 & 96.65 & \textbf{96.09} & 95.96 & 96.00 & 95.59 \\
\midrule
\multicolumn{22}{@{}l}{\textbf{5 bits} \footnotesize(5.5 bpw incl.\ scale)} \\
 uniform & RTN & \textbf{97.28} & 96.96 & 97.07 & 96.46 & \textbf{97.60} & 97.31 & 97.41 & 96.93 & \textbf{97.47} & 97.28 & 97.31 & 96.89 & \textbf{98.21} & 98.10 & 98.12 & 97.89 & \textbf{97.20} & 97.08 & 97.16 & 96.84 \\
 uniform & GPTQ & \textbf{97.63} & 96.94 & 97.19 & 96.31 & \textbf{97.95} & 97.83 & 97.88 & 97.60 & \textbf{97.77} & 97.31 & 97.46 & 96.94 & \textbf{98.42} & 98.38 & 98.37 & 98.24 & 97.42 & 97.40 & \textbf{97.43} & 97.24 \\
\midrule
\multicolumn{22}{@{}l}{\textbf{6 bits} \footnotesize(6.5 bpw incl.\ scale)} \\
 uniform & RTN & \textbf{98.32} & 98.12 & 98.21 & 98.17 & \textbf{98.48} & 98.28 & 98.37 & 98.33 & \textbf{98.43} & 98.26 & 98.34 & 98.32 & 98.83 & 98.71 & 98.78 & \textbf{98.84} & \textbf{97.85} & 97.75 & 97.78 & 97.80 \\
 uniform & GPTQ & \textbf{98.53} & 98.15 & 98.34 & 98.12 & \textbf{98.68} & 98.57 & 98.61 & 98.65 & \textbf{98.60} & 98.40 & 98.48 & 98.26 & 98.96 & 98.87 & 98.93 & \textbf{99.00} & 97.97 & 97.91 & 97.94 & \textbf{97.98} \\
\bottomrule
\end{tabular}}}
\end{table*}

Table~\ref{tab:main-ear} gives the top-10 expected acceptance rate to the BF16 model (Section~\ref{sec:method-eval}) of the models of Table~\ref{tab:main-kl}, on the same 500K held-out tokens.  Each EAR was computed on the weights of the corresponding cell of Table~\ref{tab:main-kl}, on L40S, except for RTN on Qwen3.8-27B and RTN and GPTQ on Qwen3.6-35B-A3B at 4, 5 and 6 bits (RTX A6000), and Llama-3.2-3B-Inst.\ (RTX 3090).
\section{Run-to-Run Variation}\label{app:xgpu}

\begin{table}[!htb]
\centering
\caption{One GPTQ configuration (Qwen3-8B, SH at 3 bits) from four runs that differ only in GPU type or
lazy-batch width (rows), scored on three GPU types: KL (mn).  Last column: weights that differ from the H200 run.}
\label{tab:xgpu}
\resizebox{\columnwidth}{!}{%
\begin{tabular}{@{}lrrrr@{}}
\toprule
GPTQ run & H200 & L40S & RTX A6000 & differ \\
\midrule
H200 & 62.65 & 62.69 & 62.92 & -- \\
H200, lazy batch 64 & 61.89 & 61.87 & 61.97 & 41\% \\
L40S & 66.03 & 66.12 & 66.16 & 42\% \\
RTX A6000 & 61.84 & 61.88 & 61.90 & 42\% \\
\bottomrule
\end{tabular}}
\end{table}

We isolate the effects of evaluation and quantization on the KL divergence for a single configuration: Qwen3-8B with
SH at 3 bits and GPTQ, corresponding to the cell in Table~\ref{tab:main-kl}, which reports its RTX A6000 run.
Table~\ref{tab:xgpu} reports four GPTQ runs that differ only in GPU type or lazy-batch width; each run is evaluated on
H200, L40S, and RTX A6000 GPUs.

\paragraph{Evaluation.} For this configuration, the choice of evaluation GPU has a negligible effect: when scored
across the three GPU types, identical weights exhibit KL differences of less than 0.5\%.

\paragraph{Quantization.} GPTQ is deterministic for a fixed GPU type but sensitive to the ordering of floating-point
operations. A rerun with identical code, inputs, and GPU type reproduces the result bit-for-bit; an L40S run repeated
after one week matches in every weight. Conversely, a rerun on a different GPU type, or on the H200 with half the
lazy-batch width (which only reorders floating-point sums), alters 41--42\% of the weights. Relative to the H200 run
on the same evaluation GPU, these runs exhibit KL values that are 1.2--1.6\% lower, with one exception: the L40S run
is 5.1--5.5\% higher, though its 95\% confidence interval includes zero. A single evaluation window accounts for this
entire increase; excluding it, the L40S run is 1.4\% lower. The median per-window ratio of every run to the H200 run
is 0.98--0.99. None of these differences satisfy both criteria defined in Section~\ref{sec:method-eval}: a magnitude
of at least 3\% and a 95\% confidence interval that excludes zero. This variation motivates the two criteria applied
throughout Section~\ref{sec:results}.
\begin{table*}[t]
\centering
\caption{SH parameters $(\alpha, c)$ per model and code width, chosen by the format scan on calibration KL.  $^\dagger$: LUT placement factor $2^{1/4}$; $^*$: $2^{1/2}$; $^\ddagger$: $2^{3/4}$ (otherwise 1).}
\label{tab:sh-params}
{\small\setlength{\tabcolsep}{6pt}%
\begin{tabular}{@{}llrrrrr@{}}
\toprule
 & & 2 bits & 3 bits & 4 bits & 5 bits & 6 bits \\
\midrule
Llama-3.2-3B-Inst. & $\alpha$ & 1.9375 & 1.875 & 1.9375 & 1.5$^*$ & 2.375 \\
 & $c$ & $-0.46875$ & $-0.1875$ & $-0.078125$ & $-0.125$ & $-0.09375$ \\
\addlinespace
Qwen3-8B & $\alpha$ & 0.625 & 1.75 & 1.625$^\dagger$ & 1.5$^\dagger$ & 2.25 \\
 & $c$ & $-0.15625$ & $-0.1875$ & $-0.078125$ & $-0.03125$ & $-0.125$ \\
\addlinespace
Llama-3.1-8B-Inst. & $\alpha$ & 1.375 & 1.875 & 1.625$^\ddagger$ & 1.75$^\ddagger$ & 2.5625 \\
 & $c$ & $-0.359375$ & $-0.171875$ & $-0.0625$ & $-0.09375$ & $-0.109375$ \\
\addlinespace
Qwen3.8-27B & $\alpha$ & 1.5625 & 1.4375 & 1.5 & 1.5 & 2.5625 \\
 & $c$ & $-0.390625$ & $-0.109375$ & $-0.046875$ & $-0.03125$ & $-0.015625$ \\
\addlinespace
Qwen3.6-35B-A3B & $\alpha$ & 1.5625 & 1.875 & 1.9375 & 2.25 & 2.375 \\
 & $c$ & $-0.390625$ & $-0.171875$ & $-0.078125$ & $-0.1875$ & $-0.09375$ \\
\bottomrule
\end{tabular}}
\end{table*}

\section{SH Parameters}\label{app:sh-params}

Table~\ref{tab:sh-params} lists the SH parameters $(\alpha, c)$ used in Tables~\ref{tab:main-ear} and~\ref{tab:main-kl}, with one pair per model and code width. Each pair was selected via a format scan on the calibration KL of the RTN model (Section~\ref{sec:method-scans}). The selected values vary across models and widths. For every model, $\alpha$ is largest at 6 bits (2.25--2.56), and the magnitude of the offset is smaller at 4 bits (0.047--0.078) than at 2 and 3 bits (0.11--0.47).

\section{Learned Codes}\label{app:gsq}

We test whether the advantage of SH persists when codes are learned. GSQ~\cite{dadgarnia2026gsq} extends GPTQ by learning, via Gumbel-softmax on the output error of each decoder block, a shift of up to two levels for each weight's code, along with continuous per-group scales. We added a lookup-table path to its quantizer and ran a modified version of its recipe (1024 instead of 4096 sequences, 10 instead of 20 epochs, groups of 16 instead of 128, FineWeb-Edu only) on the MLP layers of Qwen3-8B at 3 bits using the INT3, NF3, and SH3$\langle1.875,-0.171875\rangle$ codebooks. KL was scored with our protocol on 262K held-out FineWeb-Edu tokens (single runs). This pipeline differs from ours in several respects: it quantizes only the MLP layers, uses SH3$\langle1.875,-0.171875\rangle$ instead of the scanned $\langle1.75,-0.1875\rangle$, searches for BF16 scales at 1 bit per weight rather than E4M3 scales at 0.5 bits, uses unrounded tables and a different integer grid, and employs different calibration and evaluation text.

\paragraph{Code learning.}
GSQ reduces the KL of every grid by 13.4\% (INT3), 18.2\% (NF3), and 21.3\% (SH3). After training, SH3 achieves the lowest KL, which is 8.2\% below INT3 and 5.4\% below NF3; both differences have 95\% window intervals that exclude zero.

\paragraph{Size of the grid advantage.}
In this pipeline, the advantage of the grid is smaller than in our format, and learned codes contribute more to performance than the choice of grid. At GSQ's GPTQ initialization with continuous scales, the three grids lie within 2.7\% of each other, below the 3\% threshold. The margins after code learning are smaller than the 8.4\% and 24.1\% advantages by which SH leads NF and INT in the GPTQ models of Table~\ref{tab:main-kl} on the same model and width. Furthermore, the gain from code learning on INT3 exceeds the final lead of SH3 over INT3. This is consistent with part of SH's advantage compensating for the coarse E4M3 block scale; continuous scales provide this compensation by other means. We did not isolate which specific differences between the two pipelines narrow the gap between grids.
\begin{table*}[t]
\centering
\caption{Increase of the held-out KL (\%) when the block scale has no sign bit, with each grid's format re-scanned without it.}
\label{tab:flip}
{\small\setlength{\tabcolsep}{5pt}%
\begin{tabular}{@{}llrrrrrrrr@{}}
\toprule
 & & \multicolumn{4}{c}{RTN} & \multicolumn{4}{c}{GPTQ} \\
\cmidrule(lr){3-6}\cmidrule(l){7-10}
 & & SH & NF & BOF & INT & SH & NF & BOF & INT \\
\midrule
Qwen3-8B & 2 bits & +81.7 & +1947 & +1815 & +3175 & +48.8 & +37.3 & +40.7 & +52.6 \\
 & 3 bits & +32.0 & +17.7 & +25.8 & +23.1 & +20.7 & +22.3 & +23.6 & +27.0 \\
 & 4 bits & +29.3 & +16.6 & +12.5 & +10.1 & +23.2 & +17.3 & +11.7 & +13.4 \\
\midrule
Llama-3.1-8B-Inst. & 2 bits & +123 & +890 & +898 & +1195 & +102 & +42.9 & +69.6 & +94.2 \\
 & 3 bits & +34.0 & +20.3 & +19.3 & +31.0 & +37.9 & +17.3 & +23.2 & +24.4 \\
 & 4 bits & +26.9 & +19.3 & +16.5 & +11.6 & +30.9 & +16.0 & +16.4 & +12.3 \\
\bottomrule
\end{tabular}}
\end{table*}

\section{Sign of the Block Scale}\label{app:flip}

Table~\ref{tab:flip} removes the sign of the E4M3 block scale, so that every block uses the grid $G$ and never its reflection $-G$ (Section~\ref{sec:method-formats}), and repeats the format scan without it, on the two 8B models at 2 to 4 bits.  BOFS has no unsigned form.  RTN ran on RTX A6000 and GPTQ on L40S; the signed references are the configurations of Table~\ref{tab:main-kl}; BOF uses the same protocol, and Qwen3-8B GPTQ was rerun on L40S.  Without the sign, the held-out KL rises in every configuration.  At 2 bits under RTN, it rises by 890--3175\% for NF, BOF and INT and by 82--123\% for SH, whose scan moves to a larger $\alpha$.  At 3 and 4 bits, the KL of SH, whose offset relies on the sign, rises by 21--38\%, and that of NF, BOF and INT by 10--31\%.  Without the sign, SH remains below the best other grid in all 8 settings at 3 and 4 bits, by more than 3\% in 5.
\section{Grid Families}\label{app:families}

We investigate whether the performance of SH depends on the $\sinh$ form of its level density or on its two degrees of freedom, namely the level spacing and the offset (Section~\ref{sec:theory}). On Qwen3-8B and Llama-3.1-8B-Instruct, we compare SH with four additional families. NFO is NormalFloat with a free outer quantile $z$, which determines the grid width; NFOC adds an SH-style offset $c$ to NFO. GC is a compander with level density $(1+y^2\sinh^2\alpha)^{-\beta}$, which reduces to SH for $\beta=0.5$ and to the arctan compander for $\beta=1$; we evaluate $\beta=1$ and the heavier-tailed case $\beta=0.25$. DDW is a Lloyd codebook learned per layer with the lever in the loop; DDWG learns a single codebook for the entire model, while DDWU learns per-layer codebooks without channel-norm weights. Each parametric family is optimized via coordinate descent initialized at the SH parameters (for NFO, this involves a line scan over $z$). The optimization criterion is the calibration KL from an earlier scan campaign: RTN with the lever and LUT placement factor 1, where the KL is coarsened to the 128 most likely reference tokens across 64 calibration windows of 2048 tokens each, without a BOS token per window. Consequently, these values are not comparable to the held-out KL reported in Table~\ref{tab:main-kl}. Table~\ref{tab:families-calib} reports the calibration KL of the best grid for each family at 3 and 4 bits, while Table~\ref{tab:families-gptq} reports the held-out KL after GPTQ at 3 bits, both relative to SH. The scans were executed on L40S and RTX 3090 GPUs and terminated before all planned stages completed; GPTQ results are available only for 3-bit quantization.

\paragraph{Calibration KL.}
Table~\ref{tab:families-calib} reveals two primary patterns. First, the offset accounts for most of the performance gap between NF and SH. Allowing the width of NF to vary reduces its excess KL over SH only from 15.0--32.2\% to 13.0--25.9\% (NFO), whereas adding the offset closes 74--97\% of the remaining gap, leaving NFOC within 0.4--5.0\% of SH. Second, among families incorporating an offset, the specific shape of the level density has a comparatively minor impact: the arctan and heavier-tailed companders exhibit KL values 1.6--2.6\% higher than SH at 3 bits and 5.5--6.8\% higher at 4 bits. Only the per-layer Lloyd codebook matches SH across all four settings ($-1.2\%$ to $+1.1\%$); a single global codebook (DDWG) or a per-layer codebook without channel-norm weights (DDWU) yields KL values 0.3--5.9\% higher than SH.

\paragraph{After GPTQ.}
Table~\ref{tab:families-gptq} repeats the comparison at 3 bits following GPTQ, using single runs on one RTX 3090 node, each compared against an SH control run on the same hardware. The patterns observed in calibration KL persist. NF without an offset remains 8.9--16.7\% above the control, whereas NFOC and the arctan compander remain within 3.0\%, a margin consistent with GPTQ run-to-run variation (Appendix~\ref{app:xgpu}); the heavier-tailed compander lies 7.7\% above the control on Qwen3-8B and 1.9\% on Llama-3.1-8B-Instruct. Among other families, only the per-layer Lloyd codebook falls more than 3\% below the control, achieving a 6.5\% reduction on Llama-3.1-8B-Instruct with lower KL in 186 of 245 windows; for this model, the SH grid selected for Qwen3-8B also yields a 3.7\% improvement over the control. Thus, SH is not optimal in an absolute sense. Its advantage over fixed grids stems from its two degrees of freedom, which NFOC and the arctan compander replicate within run-to-run variation.

\begin{table}[t]
\centering
\caption{Calibration KL of the best grid of each family with RTN, in \% relative to SH (negative: lower than SH; --: not scanned).}
\label{tab:families-calib}
\resizebox{\columnwidth}{!}{%
\begin{tabular}{@{}lrrrr@{}}
\toprule
 & \multicolumn{2}{c}{Qwen3-8B} & \multicolumn{2}{c}{Llama-3.1-8B-Inst.} \\
\cmidrule(lr){2-3}\cmidrule(l){4-5}
 & 3 bits & 4 bits & 3 bits & 4 bits \\
\midrule
\multicolumn{5}{@{}l}{\emph{Free width and offset}} \\
NF, free width and offset (NFOC) & +0.4 & +5.0 & +4.8 & +4.9 \\
Arctan compander (GC, $\beta{=}1$) & +1.6 & +5.5 & +2.3 & +6.8 \\
Heavier-tail compander (GC, $\beta{=}0.25$) & +2.6 & -- & +2.4 & +6.4 \\
\multicolumn{5}{@{}l}{\emph{No offset}} \\
NF, free width (NFO) & +13.0 & +19.3 & +25.9 & +25.1 \\
NF & +15.0 & +23.2 & +27.2 & +32.2 \\
\multicolumn{5}{@{}l}{\emph{Learned codebooks}} \\
Lloyd per layer (DDW) & $-$1.2 & +1.1 & $-$0.1 & +0.8 \\
Lloyd per model (DDWG) & +0.8 & +0.3 & +5.9 & +0.6 \\
Lloyd per layer, unweighted (DDWU) & +2.6 & +2.2 & +5.2 & +2.2 \\
\bottomrule
\end{tabular}}
\end{table}

\begin{table}[t]
\centering
\caption{Held-out KL of the best 3-bit grid of each family after GPTQ, in \% relative to SH on the same GPU (single runs).}
\label{tab:families-gptq}
\resizebox{\columnwidth}{!}{%
\begin{tabular}{@{}lrr@{}}
\toprule
 & Qwen3-8B & Llama-3.1-8B-Inst. \\
\midrule
SH grid of the other model & $-$0.7 & $-$3.7 \\
NF, free width and offset (NFOC) & +2.3 & +0.2 \\
Arctan compander (GC, $\beta{=}1$) & +3.0 & +1.6 \\
Heavier-tail compander (GC, $\beta{=}0.25$) & +7.7 & +1.9 \\
NF, free width (NFO) & +10.4 & +16.7 \\
NF (placement $2^{3/4}$) & +8.9 & +14.0 \\
Lloyd per layer (DDW) & $-$1.2 & $-$6.5 \\
Lloyd per model (DDWG) & +2.8 & +5.2 \\
Lloyd per layer, unweighted (DDWU) & +1.5 & +1.5 \\
\bottomrule
\end{tabular}}
\end{table}
\section{Downstream Benchmarks}\label{app:bench}

We evaluate quantized models on the benchmark suite defined by BOF4~\cite{blumenberg2025bof4}: MMLU~\cite{hendrycks2021mmlu} with five shots, and ARC-Challenge~\cite{clark2018arc}, HellaSwag~\cite{zellers2019hellaswag}, PIQA~\cite{bisk2020piqa}, SIQA~\cite{sap2019siqa}, and WinoGrande~\cite{sakaguchi2020winogrande} zero-shot. All tasks follow the definitions in lm-evaluation-harness 0.4.11~\cite{biderman2024harness}, executed on its Hugging Face backend for the 8B models and on vLLM 0.32 for Qwen3.6-35B-A3B. Following BOF4, we report plain accuracy rather than length-normalized accuracy.

The RTN and GPTQ rows correspond to runs detailed in Tables~\ref{tab:main-kl} and~\ref{tab:nvfp4}; BOF, which is not listed in these tables, derives from runs under the same protocol. For the 8B models, a difference is deemed significant if its 95\% confidence interval, derived from the normal approximation of per-question differences, excludes zero. For Qwen3.6-35B-A3B, we apply a paired per-question sign-flip test to the difference in average accuracy, using the normal approximation and Holm correction across the 18 comparisons for this model.

\paragraph{Comparison with NVFP4.}
At 4.5 bpw (Table~\ref{tab:bench-nvfp4}), SH4 achieves higher average accuracy than NVFP4 in all six model-quantizer combinations, by 0.1--0.7 points. On the two 8B models, SH4 also achieves higher MMLU accuracy; all four MMLU differences are statistically significant, while the differences in average accuracy are significant only for Llama-3.1-8B-Instruct. On Qwen3.6-35B-A3B, SH4 exceeds NVFP4 in average accuracy by 0.1 points under RTN and 0.4 points after GPTQ, neither of which is significant.

\paragraph{Grids on the 8B models.}
At 3 bits, SH achieves the highest average accuracy among the five grids on both 8B models under both quantizers (Tables~\ref{tab:bench-q8b} and~\ref{tab:bench-l31}); in each of these four cases, the gap to the next-best grid is not statistically significant. At 2 bits, performance diverges. On Llama-3.1-8B-Instruct, SH leads all other grids under both quantizers, and all such differences are significant. On Qwen3-8B, BOFS exceeds SH by 0.5 points under RTN and 0.6 points after GPTQ, and INT exceeds SH by 0.2 points under RTN; none of these advantages are statistically significant.

\paragraph{Mixture-of-experts model.}
On Qwen3.6-35B-A3B (Table~\ref{tab:bench-q36}), SH achieves the lowest KL in all four settings. At 3 bits, BOFS and BOF exceed SH in average accuracy under both quantizers, though none of these 3-bit differences are significant. At 2 bits, SH achieves the highest average accuracy under RTN. After GPTQ, BOFS and INT exceed SH by 1.3 and 1.2 points, respectively, despite having higher KL values. The 2-bit differences between SH and BOF, NF, and INT under RTN, and between SH and BOFS and INT after GPTQ, are statistically significant (adjusted $p<0.05$). Across the three models, this is the only setting in which a grid with higher KL than SH achieves significantly higher accuracy.

\begin{table*}[t]
\centering
\caption{Benchmark accuracy (\%) of SH4 and NVFP4 at 4.5 bpw, with the KL (mn) and EAR (\%) of the same weights.  Avg: mean of the six tasks.  Bold: higher Avg of each pair.}
\label{tab:bench-nvfp4}
{\small\setlength{\tabcolsep}{4pt}%
\begin{tabular}{@{}lllrrrrrrrrr@{}}
\toprule
 & & & MMLU & ARC-C & HellaSwag & PIQA & SIQA & WinoGrande & Avg & KL (mn) & EAR (\%) \\
\midrule
Qwen3-8B & & BF16 & 74.9 & 55.5 & 57.2 & 76.6 & 51.6 & 68.0 & 63.9 & -- & -- \\
 & RTN & SH4 & 74.7 & 55.5 & 56.6 & 76.6 & 51.1 & 68.1 & \textbf{63.8} & 18.35 & 95.5 \\
 &  & NVFP4 & 73.6 & 55.5 & 56.3 & 76.4 & 49.9 & 69.8 & 63.6 & 48.70 & 92.5 \\
 & GPTQ & SH4 & 74.6 & 54.8 & 56.7 & 76.4 & 51.5 & 68.0 & \textbf{63.7} & 14.46 & 96.2 \\
 &  & NVFP4 & 74.1 & 54.9 & 56.4 & 76.4 & 50.8 & 67.0 & 63.3 & 32.65 & 94.3 \\
\midrule
Llama-3.1-8B-Inst. & & BF16 & 68.3 & 53.9 & 59.8 & 80.1 & 49.8 & 73.7 & 64.3 & -- & -- \\
 & RTN & SH4 & 67.7 & 53.4 & 59.4 & 80.1 & 50.0 & 74.1 & \textbf{64.1} & 21.21 & 95.2 \\
 &  & NVFP4 & 66.2 & 52.2 & 59.1 & 79.4 & 50.9 & 73.0 & 63.5 & 52.26 & 92.3 \\
 & GPTQ & SH4 & 67.7 & 52.0 & 59.4 & 80.6 & 49.9 & 74.0 & \textbf{64.0} & 19.81 & 95.7 \\
 &  & NVFP4 & 66.9 & 52.4 & 59.3 & 79.0 & 50.5 & 72.6 & 63.4 & 52.91 & 92.8 \\
\midrule
Qwen3.6-35B-A3B & & BF16 & 82.5 & 54.9 & 63.3 & 82.3 & 54.2 & 73.8 & 68.5 & -- & -- \\
 & RTN & SH4 & 82.4 & 55.2 & 63.2 & 82.6 & 54.6 & 73.7 & \textbf{68.6} & 16.68 & 95.7 \\
 &  & NVFP4 & 82.5 & 54.1 & 63.2 & 82.3 & 54.1 & 74.9 & 68.5 & 36.64 & 93.5 \\
 & GPTQ & SH4 & 82.7 & 54.9 & 63.4 & 82.4 & 54.0 & 73.4 & \textbf{68.5} & 14.44 & 96.1 \\
 &  & NVFP4 & 82.2 & 54.0 & 63.1 & 82.3 & 52.6 & 74.4 & 68.1 & 24.35 & 94.9 \\
\bottomrule
\end{tabular}
}
\end{table*}

\begin{table*}[t]
\centering
\caption{Benchmark accuracy (\%) at 2 and 3 bits on Qwen3-8B, with the KL (mn) and EAR (\%) of the same weights.  Bold: best per column within each block (lowest KL, highest accuracy and EAR).}
\label{tab:bench-q8b}
{\small\setlength{\tabcolsep}{4pt}%
\begin{tabular}{@{}llrrrrrrrrr@{}}
\toprule
 & & MMLU & ARC-C & HellaSwag & PIQA & SIQA & WinoGrande & Avg & KL (mn) & EAR (\%) \\
\midrule
\multicolumn{2}{@{}l}{BF16} & 74.9 & 55.5 & 57.2 & 76.6 & 51.6 & 68.0 & 63.9 & -- & -- \\
\midrule
\multicolumn{11}{@{}l}{\textbf{3 bits} \footnotesize(3.5 bpw incl.\ scale)} \\
RTN & SH & \textbf{73.7} & 52.9 & \textbf{55.5} & 75.9 & 48.8 & \textbf{69.1} & \textbf{62.6} & \textbf{78.31} & \textbf{90.7} \\
 & BOFS & 72.7 & \textbf{53.8} & 55.4 & 75.7 & 48.4 & 68.2 & 62.4 & 87.23 & 90.0 \\
 & BOF & 72.4 & 53.5 & 54.9 & \textbf{76.7} & 48.7 & 68.4 & 62.4 & 86.46 & -- \\
 & NF & 73.4 & 53.5 & 55.3 & 75.5 & \textbf{49.7} & 67.7 & 62.5 & 88.60 & 90.1 \\
 & INT & 71.8 & 53.2 & 54.9 & 76.2 & 48.8 & 66.3 & 61.9 & 109.6 & 88.7 \\
\addlinespace
GPTQ & SH & 72.9 & \textbf{56.7} & 55.2 & 75.8 & 49.0 & \textbf{68.7} & \textbf{63.0} & \textbf{61.90} & \textbf{92.0} \\
 & BOFS & \textbf{73.1} & 54.5 & \textbf{55.3} & 76.1 & 49.7 & 67.9 & 62.8 & 66.94 & 91.7 \\
 & BOF & 72.9 & 53.9 & \textbf{55.3} & \textbf{76.6} & 49.3 & 68.0 & 62.7 & 67.36 & -- \\
 & NF & 72.4 & 53.1 & 55.2 & 76.0 & 49.6 & 68.0 & 62.4 & 67.60 & 91.6 \\
 & INT & 72.4 & 53.9 & 54.8 & 76.2 & \textbf{50.3} & 67.8 & 62.6 & 81.51 & 90.8 \\
\midrule
\multicolumn{11}{@{}l}{\textbf{2 bits} \footnotesize(2.5 bpw incl.\ scale)} \\
RTN & SH & 62.0 & 43.1 & 45.5 & 71.9 & 45.4 & 61.8 & 55.0 & \textbf{520.6} & \textbf{75.5} \\
 & BOFS & \textbf{63.3} & 42.9 & \textbf{45.7} & 72.4 & 45.2 & \textbf{63.1} & \textbf{55.4} & 561.8 & 74.7 \\
 & BOF & 57.0 & 39.4 & 44.8 & 71.3 & 43.0 & 62.0 & 52.9 & 642.1 & -- \\
 & NF & 57.5 & 39.6 & 45.0 & 72.0 & 42.6 & 61.6 & 53.1 & 601.8 & 73.9 \\
 & INT & 60.7 & \textbf{44.2} & 45.2 & \textbf{72.6} & \textbf{45.5} & 62.8 & 55.2 & 618.2 & 73.2 \\
\addlinespace
GPTQ & SH & \textbf{64.3} & 44.6 & 47.1 & \textbf{73.1} & 46.3 & 65.5 & 56.8 & 379.8 & \textbf{79.2} \\
 & BOFS & 64.0 & \textbf{47.4} & \textbf{47.4} & 72.6 & \textbf{47.4} & 65.7 & \textbf{57.4} & \textbf{378.2} & \textbf{79.2} \\
 & BOF & 59.7 & 43.2 & 46.5 & 71.3 & 44.3 & 67.4 & 55.4 & 425.3 & -- \\
 & NF & 60.6 & 41.9 & 46.3 & 71.3 & 44.1 & \textbf{67.7} & 55.3 & 432.5 & 77.7 \\
 & INT & 63.7 & 45.5 & 46.7 & 71.8 & 46.3 & 65.7 & 56.6 & 405.3 & 78.3 \\
\bottomrule
\end{tabular}
}
\end{table*}

\begin{table*}[t]
\centering
\caption{Benchmark accuracy (\%) at 2 and 3 bits on Llama-3.1-8B-Inst., with the KL (mn) and EAR (\%) of the same weights.  Bold: best per column within each block (lowest KL, highest accuracy and EAR).}
\label{tab:bench-l31}
{\small\setlength{\tabcolsep}{4pt}%
\begin{tabular}{@{}llrrrrrrrrr@{}}
\toprule
 & & MMLU & ARC-C & HellaSwag & PIQA & SIQA & WinoGrande & Avg & KL (mn) & EAR (\%) \\
\midrule
\multicolumn{2}{@{}l}{BF16} & 68.3 & 53.9 & 59.8 & 80.1 & 49.8 & 73.7 & 64.3 & -- & -- \\
\midrule
\multicolumn{11}{@{}l}{\textbf{3 bits} \footnotesize(3.5 bpw incl.\ scale)} \\
RTN & SH & \textbf{65.0} & \textbf{52.4} & \textbf{58.1} & \textbf{79.4} & \textbf{50.2} & 73.2 & \textbf{63.1} & \textbf{90.82} & \textbf{89.9} \\
 & BOFS & 64.7 & 51.8 & 57.8 & 78.7 & 49.6 & 73.1 & 62.6 & 107.0 & 89.0 \\
 & BOF & 64.4 & 51.3 & 58.0 & 78.9 & 50.1 & 73.2 & 62.7 & 109.4 & -- \\
 & NF & 64.7 & 51.5 & 58.0 & 78.7 & 49.7 & 72.1 & 62.5 & 105.2 & 89.1 \\
 & INT & 64.3 & 49.7 & 57.5 & 78.9 & 49.3 & \textbf{73.5} & 62.2 & 138.0 & 87.4 \\
\addlinespace
GPTQ & SH & \textbf{66.5} & 50.4 & 57.8 & \textbf{79.4} & \textbf{50.1} & 72.9 & \textbf{62.9} & \textbf{81.46} & \textbf{90.9} \\
 & BOFS & 65.4 & 50.9 & \textbf{58.2} & 79.2 & 47.1 & 72.5 & 62.2 & 98.25 & 90.0 \\
 & BOF & 65.6 & \textbf{52.8} & 58.0 & 78.9 & 48.9 & 72.8 & 62.8 & 95.25 & -- \\
 & NF & 65.5 & 50.2 & 57.8 & 78.8 & 48.5 & \textbf{73.1} & 62.3 & 95.94 & 90.2 \\
 & INT & 65.0 & 51.5 & 57.6 & 79.3 & 48.7 & 73.0 & 62.5 & 118.8 & 88.9 \\
\midrule
\multicolumn{11}{@{}l}{\textbf{2 bits} \footnotesize(2.5 bpw incl.\ scale)} \\
RTN & SH & \textbf{47.0} & 40.5 & \textbf{48.6} & \textbf{75.2} & 45.0 & \textbf{68.1} & \textbf{54.1} & \textbf{643.2} & \textbf{73.4} \\
 & BOFS & 41.9 & \textbf{40.6} & 46.5 & 74.0 & \textbf{46.4} & 65.5 & 52.5 & 819.0 & 70.6 \\
 & BOF & 39.8 & 33.7 & 43.1 & 71.8 & 43.7 & 65.5 & 49.6 & 1102.2 & -- \\
 & NF & 40.8 & 32.8 & 44.5 & 71.6 & 42.9 & 65.8 & 49.7 & 1001.7 & 67.5 \\
 & INT & 40.9 & 39.2 & 44.9 & 72.7 & 45.1 & 65.0 & 51.3 & 955.4 & 67.9 \\
\addlinespace
GPTQ & SH & \textbf{53.5} & 40.4 & \textbf{48.8} & \textbf{75.5} & \textbf{46.7} & \textbf{69.6} & \textbf{55.8} & \textbf{509.0} & \textbf{76.6} \\
 & BOFS & 50.6 & \textbf{41.5} & 46.7 & 73.4 & 44.2 & 69.5 & 54.3 & 628.8 & 74.4 \\
 & BOF & 45.2 & 37.6 & 45.3 & 72.2 & 44.3 & 66.8 & 51.9 & 699.8 & -- \\
 & NF & 46.1 & 37.6 & 45.0 & 71.9 & 45.0 & 67.3 & 52.2 & 727.8 & 72.5 \\
 & INT & 48.5 & 39.5 & 44.7 & 72.0 & 42.3 & 68.8 & 52.6 & 685.7 & 73.3 \\
\bottomrule
\end{tabular}
}
\end{table*}

\begin{table*}[t]
\centering
\caption{Benchmark accuracy (\%) at 2 and 3 bits on Qwen3.6-35B-A3B, with the KL (mn) and EAR (\%) of the same weights.  Bold: best per column within each block (lowest KL, highest accuracy and EAR).}
\label{tab:bench-q36}
{\small\setlength{\tabcolsep}{4pt}%
\begin{tabular}{@{}llrrrrrrrrr@{}}
\toprule
 & & MMLU & ARC-C & HellaSwag & PIQA & SIQA & WinoGrande & Avg & KL (mn) & EAR (\%) \\
\midrule
\multicolumn{2}{@{}l}{BF16} & 82.5 & 54.9 & 63.3 & 82.3 & 54.2 & 73.8 & 68.5 & -- & -- \\
\midrule
\multicolumn{11}{@{}l}{\textbf{3 bits} \footnotesize(3.5 bpw incl.\ scale)} \\
RTN & SH & 83.3 & 56.4 & 62.6 & 81.9 & 53.2 & 72.8 & 68.3 & \textbf{55.40} & \textbf{92.0} \\
 & BOFS & 83.6 & \textbf{57.6} & 62.4 & \textbf{82.3} & 53.3 & \textbf{74.2} & \textbf{68.9} & 63.74 & 91.3 \\
 & BOF & \textbf{84.1} & 56.9 & 62.7 & 81.7 & \textbf{54.4} & 73.4 & \textbf{68.9} & 63.35 & -- \\
 & NF & 83.7 & 56.0 & \textbf{62.8} & 81.5 & 52.4 & 73.2 & 68.2 & 62.45 & 91.5 \\
 & INT & 83.7 & 56.2 & 62.4 & 82.2 & 53.0 & 73.2 & 68.5 & 77.77 & 90.4 \\
\addlinespace
GPTQ & SH & \textbf{84.2} & 54.9 & \textbf{63.0} & 81.2 & 52.4 & 73.3 & 68.1 & \textbf{46.92} & \textbf{92.8} \\
 & BOFS & \textbf{84.2} & 55.0 & 62.7 & 81.9 & \textbf{54.6} & 72.8 & 68.5 & 49.45 & 92.6 \\
 & BOF & 84.0 & 55.4 & 62.7 & 82.0 & 53.8 & \textbf{73.4} & \textbf{68.6} & 49.27 & -- \\
 & NF & 83.7 & \textbf{56.3} & 62.7 & 82.2 & 51.6 & 72.8 & 68.2 & 49.63 & 92.6 \\
 & INT & 84.0 & 55.0 & 62.7 & \textbf{82.5} & 51.1 & 72.1 & 67.9 & 57.83 & 92.0 \\
\midrule
\multicolumn{11}{@{}l}{\textbf{2 bits} \footnotesize(2.5 bpw incl.\ scale)} \\
RTN & SH & \textbf{78.8} & \textbf{50.9} & \textbf{58.3} & 79.5 & 46.6 & \textbf{71.3} & \textbf{64.2} & \textbf{250.6} & \textbf{82.8} \\
 & BOFS & 77.7 & 50.0 & 58.2 & \textbf{80.2} & 48.2 & 70.6 & 64.1 & 289.7 & 81.5 \\
 & BOF & 76.3 & 49.6 & 56.3 & 79.1 & 47.5 & 69.9 & 63.1 & 342.6 & -- \\
 & NF & 74.8 & 49.6 & 56.3 & 79.1 & 45.0 & 68.0 & 62.1 & 336.7 & 80.1 \\
 & INT & 75.7 & 49.7 & 56.4 & 78.6 & \textbf{49.6} & 69.3 & 63.2 & 328.7 & 80.2 \\
\addlinespace
GPTQ & SH & \textbf{79.1} & 52.0 & \textbf{58.9} & 79.7 & 45.8 & 70.5 & 64.3 & \textbf{208.2} & \textbf{84.5} \\
 & BOFS & 77.9 & 55.5 & 58.7 & \textbf{80.7} & \textbf{50.1} & 70.7 & \textbf{65.6} & 224.4 & 83.8 \\
 & BOF & 77.6 & 53.1 & 57.3 & 78.5 & 47.2 & 70.9 & 64.1 & 258.2 & -- \\
 & NF & 76.8 & 53.3 & 57.4 & 79.4 & 47.3 & 70.0 & 64.1 & 263.6 & 82.4 \\
 & INT & 78.2 & \textbf{55.7} & 58.1 & 80.3 & 48.6 & \textbf{72.2} & 65.5 & 240.5 & 83.1 \\
\bottomrule
\end{tabular}
}
\end{table*}
\section{Per-Layer Grids}\label{app:per-layer}
\begin{figure}[!htb]
\centering
\includegraphics{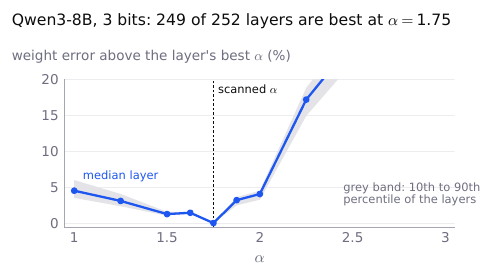}
\caption{RTN weight error of the 252 linear layers of Qwen3-8B at 3 bits against $\alpha$ ($c=-0.1875$), in \% above
each layer's lowest error over the ten values.  Dashed: the $\alpha$ chosen by the format scan.}
\label{fig:sh-layers}
\end{figure}
Tuning per-layer grids around global choices rarely reduces KL after GPTQ. For SH, NF, and the superior of BOF/BOFS at 2--6 bits on both 8B models, RCO selected the LUT placement factor for each layer; for SH, it also optimized $\alpha$ and $c$ within $\alpha\pm0.25$ and $c\pm1/64$, based on calibration KL with RTN weights. Under RTN, this reduces KL on the first 50K evaluation tokens by 11--28\% at 2 bits and 4--15\% at 3 bits. After GPTQ, 26 of 30 configurations remain within $\pm2.2\%$ of untuned grids. The only gains exceeding 3\% that persist in at least two-thirds of the 245 windows occur on Llama-3.1-8B-Instruct (4.0--6.3\%). End-to-end selection among 30 GPTQ-quantized candidates per layer for Qwen3-8B at 3 bits reduces KL by 2.5\% and 3.2\% from two starting points, a lower bound given incomplete search convergence. Under unweighted weight error, layers prefer nearly identical grids: normalized by absolute maximum, blocks of 16 weights exhibit similar mean sorted profiles across layers. With scale search excluding channel norms, the scanned $\alpha$ is the best of ten values for 249 of 252 layers in Qwen3-8B at 3 bits (Figure~\ref{fig:sh-layers}). When weighted by squared input channel norms, as in the lever, 7 of 21 sampled layers prefer a different $\alpha$. The reason RTN gains from per-layer search do not persist after GPTQ was not isolated. Thus, using a single grid per model and width incurs minimal loss. Appendix~\ref{app:gsq} compares grids when GSQ~\cite{dadgarnia2026gsq} jointly learns codes and scales.

\end{document}